\documentclass{article}
\usepackage{scalefnt,letltxmacro}
\LetLtxMacro{\oldtextsc}{\textsc}
\renewcommand{\textsc}[1]{\oldtextsc{\scalefont{1.10}#1}}

\usepackage[acronym,smallcaps]{glossaries}
\glsdisablehyper
\makeglossaries
\usepackage{xspace}
\usepackage{float}
\usepackage{amssymb}
\usepackage{mathtools}
\usepackage{amsfonts}
\usepackage{amsmath}
\usepackage{amsthm}
\usepackage{mathrsfs}

\usepackage{pifont}
\newcommand{\cmark}{ \textcolor{green!60!black}{\ding{51}} }
\newcommand{\xmark}{ \textcolor{red!60!black}{\ding{55}} }

\usepackage{tikz}
\tikzset{
     block/.style={rectangle, draw, fill=red!40, text width=6em,
                   text centered, rounded corners, minimum height=3em},
     arrow/.style={-{Stealth[]}}
     }
\usetikzlibrary{positioning,arrows.meta, shapes.geometric, fit}

\newcommand{\mathbold}[1]{\ensuremath{\boldsymbol{\mathbf{#1}}}}

\newcommand{\nestedmathbold}[1]{{\mathbold{#1}}}

\newcommand{\mbf}{\nestedmathbold{f}}

\newcommand{\mbr}{\nestedmathbold{r}}

\newcommand{\mbu}{\nestedmathbold{u}}
\newcommand{\mbv}{\nestedmathbold{v}}

\newcommand{\mbx}{\nestedmathbold{x}}

\newcommand{\mbA}{\nestedmathbold{A}}

\newcommand{\mbC}{\nestedmathbold{C}}

\newcommand{\mbG}{\nestedmathbold{G}}

\newcommand{\mbI}{\nestedmathbold{I}}

\newcommand{\mbM}{\nestedmathbold{M}}

\newcommand{\mbW}{\nestedmathbold{W}}
\newcommand{\mbX}{\nestedmathbold{X}}
\newcommand{\mbY}{\nestedmathbold{Y}}

\newcommand{\mbtheta}{\nestedmathbold{\theta}}

\newcommand{\mbxi}{\nestedmathbold{\xi}}

\newcommand{\mbSigma}{\nestedmathbold{\Sigma}}

\DeclareRobustCommand{\KL}[2]{\ensuremath{\textsc{kl}\left[#1\;\|\;#2\right]}}
\DeclareRobustCommand{\chisq}[2]{\ensuremath{\chi^2\left[#1\;\|\;#2\right]}}

\DeclarePairedDelimiterX{\infdivx}[2]{[}{]}{%
  #1\;\delimsize\|\;#2%
}

\newcommand{\dx}[1]{\text{d} #1}

\newcommand{\norm}[1]{\left\lVert#1\right\rVert}

\DeclareMathOperator*{\maxargmax}{\max\cdot\arg\max}
\newacronym{MAP}{map}{maximum-a-posteriori}
\newacronym{MLE}{mle}{maximum likelihood estimation}
\newacronym{MNLL}{mnll}{mean negative loglikelihood}

\newacronym{MCMC}{mcmc}{Markov chain Monte Carlo}
\newacronym{HMC}{hmc}{Hamiltonian Monte Carlo}
\newacronym{LD}{ld}{Langevin dynamics}
\newacronym{SGNHT}{sgnht}{stochastic gradient Nos\'e-Hoover thermostat}
\newacronym{MH}{mh}{Metropolis-Hastings}
\newacronym{NUTS}{nuts}{no-u-turn sampler}
\newacronym{SGHMC}{sghmc}{stochastic gradient Hamiltonian Monte Carlo}
\newacronym{VI}{vi}{variational inference}
\newacronym{KL}{kl}{Kullback-Leibler divergence}
\newacronym{RKHS}{rkhs}{reproducing kernel Hilbert space}
\newacronym{PO}{po}{particle optimization}
\newacronym{SVGD}{svgd}{Stein variational gradient descent}

\newacronym{SDE}{sde}{stochastic differential equation}
\newacronym{FPE}{fpe}{Fokker-Planck equation}
\newacronym{PDE}{pde}{partial differential equation}
\newacronym{SGE}{sge}{Stein gradient estimator}
\newacronym{WGF}{wgf}{Wasserstein gradient flow}
\newcommand{\ito}{It\=o}
\newacronym{KSD}{ksd}{kernelized Stein discrepancy}
\newacronym{GSVGD}{gsvgd}{generalized SVGD}
\newacronym{PARVI}{parvi}{particle-based variational inference}
\newacronym{GF}{gf}{gradient flow}

\newacronym{LAWGD}{lawgd}{Laplacian Adjusted Wasserstein Gradient Descent}
\newacronym{SGRHMC}{sgrhmc}{stochastic gradient Riemannian Hamiltonian Monte Carlo}
\newacronym{ODE}{ode}{ordinary differential equation}

\PassOptionsToPackage{round}{natbib}

\usepackage[preprint]{neurips_2021}

\usepackage[utf8]{inputenc} 
\usepackage[T1]{fontenc}    
\usepackage{url}            
\usepackage{booktabs}       
\usepackage{amsfonts}       
\usepackage{nicefrac}       
\usepackage{microtype}      
\usepackage{xcolor}         

\definecolor{linkblue}{rgb}{0.10,0.40,0.70}
\usepackage[colorlinks=true,citecolor=linkblue]{hyperref}

\title{De-randomizing MCMC dynamics with the diffusion Stein operator}

\author{%
  Zheyang Shen 
  \qquad
  Markus Heinonen 
  \qquad 
  Samuel Kaski\\
  Department of Computer Science\\
  Aalto University, Finland\\
  \texttt{first.last@aalto.fi}
}

\begin{document}

\maketitle

\begin{abstract}
Approximate Bayesian inference estimates descriptors of an intractable target distribution -- in essence, an optimization problem within a family of distributions. 
For example, \gls{LD} extracts asymptotically exact samples from a diffusion process because the time evolution of its marginal distributions constitutes a curve that minimizes the KL-divergence via steepest descent in the Wasserstein space. 
Parallel to LD, \gls{SVGD} similarly minimizes the KL 
, albeit endowed with a novel Stein-Wasserstein distance, by \emph{deterministically} transporting a set of particle samples, thus de-randomizes the stochastic diffusion process.
We propose de-randomized kernel-based particle samplers to all diffusion-based samplers known as MCMC dynamics. 
Following previous work in interpreting MCMC dynamics, we equip the Stein-Wasserstein metric with a fiber-Riemannian Poisson structure, with the capacity of characterizing a fiber-gradient Hamiltonian flow that simulates MCMC dynamics. 
Such dynamics discretize into generalized \gls{SVGD} (GSVGD), a Stein-type deterministic particle sampler, with particle updates coinciding with applying the \emph{diffusion Stein operator} to a kernel function. 
We demonstrate empirically that GSVGD can de-randomize complicated MCMC dynamics, which combine the advantages of auxiliary momentum variables and Riemannian structure, while maintaining the high sample quality from an interacting particle system. 
\end{abstract}
\section{Introduction}
Evaluating an un-normalized target distribution $\pi$ is a centerpiece of Bayesian inference, due to its ubiquitous presence in posterior distributions. \gls{MCMC} methods fulfill this objective by generating asymptotically exact random samples from the distribution, a significant subset of which involves discretization of continuous-time diffusion processes, most notably Langevin diffusion, stochastic gradient \gls{HMC} \citep{chen2014stochastic} and their further generalizations, which we collectively call \emph{\gls{MCMC} dynamics} \citep{ma2015a}. Despite its simplicity and theoretical soundness, this diffusion-based sampling often suffers from slow convergence and small effective sample sizes, largely due to the auto-correlation of the samples.

As an alternative to the simulation of stochastic systems, \acrlong{PARVI} \citep{liu2016stein, chen2018a, liu2019understanding} partially addresses the shortcomings of \gls{MCMC} by replacing the Langevin diffusion, the simplest \gls{MCMC} dynamics, with a deterministic interacting particle system that transports a set of interacting particles towards the target distribution. 
Theoretically speaking, Langevin diffusion encodes an evolution of density that minimizes the KL-divergence through steepest descent in the 2-Wasserstein space \citep{jordan1998the}, and \gls{PARVI} approximates such evolution using \gls{RKHS} \citep{liu2017stein, liu2019understanding}, thus de-randomizing the Langevin diffusion process.\par
Given the elegant theoretical link between Langevin diffusion and \gls{PARVI}, one crucial question arises: can we leverage the advantages of general \gls{MCMC} dynamics onto de-randomized particle systems? While it is difficult to formulate them as a direct Wasserstein gradient flow, \citet{liu2019mcmc} propose an interpretation for ``regular'' \gls{MCMC} dynamics that augments the 2-Wasserstein space with a fiber bundle, forming a fiber-Riemannian Poisson manifold, under which \gls{MCMC} dynamics follows a Hamiltonian flow on the fiber bundle, and a gradient flow on each fiber.\par
We show that adapting the fiber-Riemannian Hamiltonian flow to the Stein-Wasserstein metric \citep{liu2017stein,duncan2019on} yields a vector field in the form of applying the \emph{diffusion Stein operator} to the kernel function $k(\cdot, \mbtheta)$. A discretization of this vector field gives 
\gls{GSVGD}, an interacting particle system that generalizes \gls{SVGD}, with particle updates balancing an attractive force maximizing the log-likelihood and a repulsive force preventing a ``mode collapse'' of particles. We further demonstrate that the connection drawn between \gls{LD} and \gls{SVGD} \citep{liu2016stein, liu2017stein, liu2019understanding} is retraced by \gls{GSVGD}, reaffirming our claim that \gls{GSVGD} mirrors \gls{SVGD} in approximating a larger class of \gls{MCMC} diffusion processes.
Within the generality of our framework, we can develop \gls{PARVI} algorithms that exploit two key types of possible acceleration in \gls{MCMC} dynamics: auxiliary momentum variables and an adaptive Riemannian parametrization that allows for fast and efficient exploration of the probability space \citep{girolami2011riemann}, as shown in Table \ref{tab:table1}.\par
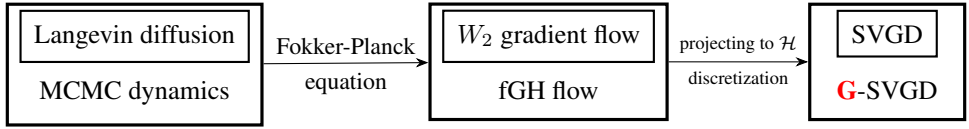
\begin{figure}
    \centering
    \begin{tikzpicture}[textnode/.style={anchor=base,inner sep=0pt}]
\node[textnode] (LD) at (0,0) {Langevin diffusion};
\node[textnode] (MCMC) at (0, -0.75) {MCMC dynamics};
\node[draw, very thick, inner xsep=1em, inner ysep=0.75em, fit=(LD) (MCMC)] (boxbig) {};
\node[draw, thick, inner xsep=0.5em, inner sep=0.5em, fit=(LD)] (boxsmall) {};

\node[textnode] (W2) at (5.5,0) {$W_2$ gradient flow};
\node[textnode] (W2AC) at (5.5, -0.75) {fGH flow};
\node[draw, very thick, inner xsep=1em, inner ysep=0.75em, fit=(W2AC) (W2)] (boxbig1) {};
\node[draw, thick, inner xsep=0.5em, inner sep=0.5em, fit=(W2)] (boxsmall1) {};

\node[textnode] (SVGD) at (10,0) {SVGD};
\node[textnode] (GSVGD) at (10, -0.75) {\textbf{{\color{red}G}}-SVGD};
\node[draw, very thick, inner xsep=1em, inner ysep=0.75em, fit=(SVGD) (GSVGD)] (boxbig2) {};
\node[draw, thick, inner xsep=0.5em, inner sep=0.5em, fit=(SVGD)] (boxsmall2) {};
\draw[arrow] (boxbig)  --  (boxbig1) node [above,pos=0.5] {\small{Fokker-Planck}} node [below,pos=0.5] {\small{equation}};
\draw[arrow] (boxbig1)  --  (boxbig2) node [above,pos=0.5] {\scriptsize{projecting to $\mathcal{H}$}} node [below,pos=0.5] {\scriptsize{discretization}};
\end{tikzpicture}
    \caption{A diagram showing the contribution of our paper: we extend and generalize previous work linking Langevin diffusion to gradient flow on the 2-Wasserstein metric $W_2$ through the \acrlong{FPE}, and the $W_2$ gradient flow linking to \gls{SVGD} \citep{liu2016stein} through projection onto an \gls{RKHS}; 
    \gls{MCMC} dynamics \citep{ma2015a} are interpreted as a fiber-gradient Hamiltonian (fGH) flow on $W_2$ \citep{liu2019mcmc}, and a projection onto an \gls{RKHS} yields \gls{GSVGD}.}
    \label{fig:fig1}
\end{figure}
\begin{table}[ht!]
    \centering
    \resizebox{\textwidth}{!}{
    \begin{tabular}{l|cc|c|c|l}
    \toprule
        MCMC dynamics & $\mbA$& $\mbC$ & auxiliary variable & Riemannian & \gls{PARVI} variant\\
        \midrule
        SGLD \citep{welling2011bayesian} & $\mbI$ & $\mathbf{0}$ & \xmark & \xmark & \begin{tabular}{@{}l@{}}
                  SVGD \citep{liu2016stein}\\
                  Blob \citep{chen2018a} \\
                 \end{tabular} \\
        SGRLD \citep{girolami2011riemann} & $\mbG(\mbtheta)^{-1}$ & $\mathbf{0}$ & \xmark & \cmark & Riemannian SVGD \citep{liu2017riemannian} \\
        SGHMC \citep{chen2014stochastic} & $\left(\begin{matrix}\mathbf{0}&\mathbf{0}\\ \mathbf{0} & A\mbI\end{matrix}\right)$ & $\left(\begin{matrix}\mathbf{0} & -\mbI \\ \mbI &\mathbf{0}\end{matrix}\right)$ & \cmark & \xmark & HMC-blob \citep{liu2019mcmc}\\
        SGRHMC \citep{ma2015a} & $\left(\begin{matrix}\mathbf{0}&\mathbf{0}\\ \mathbf{0} & \mbG(\mbtheta)^{-1}\end{matrix}\right)$& $\left(\begin{matrix}\mathbf{0} & -\mbG(\mbtheta)^{-1/2} \\ \mbG(\mbtheta)^{-1/2} &\mathbf{0}\end{matrix}\right)$& \cmark & \cmark & this work (SGRHMC-Stein) \\
        \bottomrule
    \end{tabular}
    }
    \includegraphics[width=0.75\textwidth]{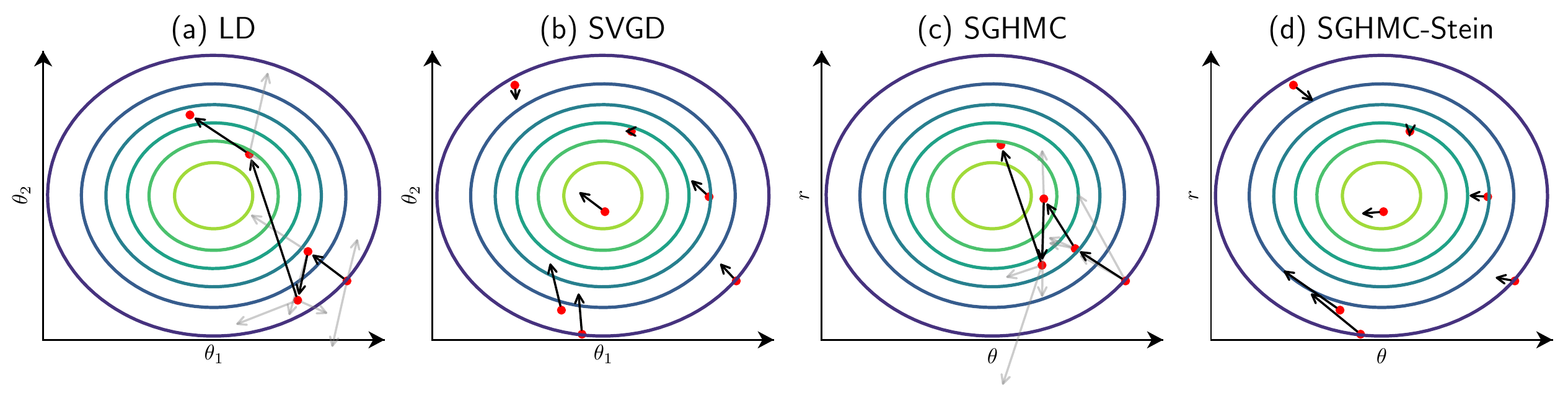}
    \caption{An overview of \gls{MCMC} dynamics along with their \gls{PARVI} approximations. Between \textbf{(a), (c)} and \textbf{(b), (d)}, \gls{MCMC} dynamics are stochastic single-chain simulations of a diffusion process (shown by one instance of a particle trajectory); \gls{PARVI} approximates the time evolution of \gls{MCMC} densities with a set of particles with deterministic dynamics; Between \textbf{(a), (b)} and \textbf{(c), (d)}, the two systems follows different time evolutions, as \textbf{(a), (b)} describe a gradient flow and \textbf{(c), (d)} describe a fiber-gradient Hamiltonian flow (with momentum variable $r$).}
    \label{tab:table1}
\end{table}

\section{MCMC dynamics -- the relevant bits}

In this paper, we consider the problem of extracting samples $\mbtheta \in \mathbb{R}^{D} = \Omega$ from an un-normalized target distribution $\pi(\mbtheta) \propto e^{-H(\mbtheta)}$. We begin by reviewing the key characteristics of \gls{MCMC} dynamics, with particular emphasis on Langevin dynamics and its \gls{FPE}, the \gls{PDE} that depicts the time evolution of a diffusion process. As the \gls{FPE} of \gls{LD} conforms to a gradient flow structure in the 2-Wasserstein metric space of probability measures $\left(\mathcal{P}(\Omega), W_2\right)$  \citep{jordan1998the}, we then briefly cover gradient flow on $\mathcal{P}(\Omega)$ as an infinite-dimensional Riemannian manifold. Through the lens of gradient flow, we can see \gls{SVGD} as a deterministic interacting particle system that approximates the gradient flow of \gls{LD} through (i) gradient flow on a novel Stein-Wasserstein metric \citep{liu2017stein}
or (ii) projecting the gradient flow direction onto an \gls{RKHS} \citep{liu2019understanding}.
With \gls{WGF} neatly linking to both \gls{LD} and \gls{SVGD}, we see that a diffusion process \gls{LD} and an interacting particle system \gls{SVGD} are, respectively, a stochastic instance and a deterministic approximation of the same Wasserstein gradient flow.

\textbf{Notations}: We use $\nabla f$ to denote the gradient of a scalar-valued function, and $\nabla\cdot \mbf$ the divergence of a vector-valued function, $\nabla\cdot\mbA$ applies the divergence operator to each row of a matrix-valued function, $\dot{\rho}_t$ the ``partial derivative'' with respect to $t$.
\subsection{\Acrlong{LD} and its \acrlong{FPE}}

In this paper, we consider \gls{MCMC} dynamics in the form of \ito\ diffusion processes following the \gls{SDE} formula
\begin{align}
    \dx{\mbtheta_t} = \mbf(\mbtheta_t)\dx{t} + \sqrt{2\mbSigma(\mbtheta_t)} \dx{\mbW_t}, \label{eq:ito}
\end{align}
consisting of drift coefficient $\mbf: \mathbb{R}^D\mapsto\mathbb{R}^D$, diffusion coefficient $\sqrt{2\mbSigma}$ and a $D$-dimensional Brownian motion $\mbW_t$. \citet{ma2015a} provide a complete recipe of all \ito\ diffusion processes converging to the target measure $\pi$.
The simplest \gls{MCMC} dynamics takes the form of Langevin diffusion \citep{langevin1908sur}, which moves towards higher densities with $\mbf(\mbtheta) = \nabla \log \pi(\mbtheta)$ perturbed by white noise $\mbSigma = \mbI$. 
Given an initial distribution $\mbtheta_0\sim\rho_0$, the \acrlong{FPE} describes the time evolution of the density of $\mbtheta_t$ \citep{risken1996fokker}
\begin{align}
    \dot{\rho}_t + \nabla\cdot \left(\rho_t \nabla\log\pi \right) - (\nabla\nabla): (\rho_t \mbI) = 0, \label{eq:fpe_ld_1}
\end{align}
where $\mbX:\mbY = \text{tr}(\mbX\top\mbY)$. Given that $(\nabla\nabla): \mbM = \nabla\cdot(\nabla\cdot\mbM)$, we can rewrite \eqref{eq:fpe_ld_1} as
\begin{align}
    \dot{\rho}_t &= \nabla\cdot \left(\nabla \rho_t-\rho_t \nabla\log\pi\right) = \nabla \cdot \bigg(\rho_t\nabla \underbrace{\frac{\delta \KL{\rho_t}{\pi}}{\delta\rho_t}}_{\text{first variation of $\KL{\rho_t}{\pi}$}}\bigg), \label{eq:fpe_ld_2}
\end{align}
a crucial step in developing the \acrlong{WGF} perspective of \gls{LD}, as $\nabla\delta E / \delta\rho$ coincides with the differential of functionals induced by the Wasserstein metric. 
\subsection{Gradient flow on $\left(\mathcal{P}(\Omega), W_2\right)$}
The conventional gradient flow (or the steepest descent curve) that minimizes a smooth function $F: \mathbb{R}^D\mapsto\mathbb{R}$ follows the \gls{PDE}: $\dot{\mbx}_t + \nabla F(\mbx_t) = 0$. Defining gradient flow on $\mathcal{P}(\Omega)$, informally written as $\dot{\rho}_t + \nabla_W E(\rho_t) = 0$, requires a definition of the differentiation $\nabla_W$, which requires an understanding of the Wasserstein metric, i.e., the inner product $g_\rho$ defined on its tangent space.\par
To simplify the discussion, we restrict the discussion on measures with a density function and with finite second-order moments. For each $\rho$, the tangent space constitutes smooth functions integrating to zero, $T_\rho\mathcal{P}(\Omega)=\{f\vert f \in C^\infty(\Omega), \int f(\mbtheta)\dx{\mbtheta} = 0\}$, in that a curve on $\mathcal{P}(\Omega)$ preserves volume. The cotangent space constitutes an equivalence class of smooth functions in differing constant, noted as the quotient space $T^*_\rho \mathcal{P}(\Omega) =C^\infty(\Omega)/\mathbb{R}$. We characterize the inner product space using the metric tensor $G(\rho): T_\rho \mathcal{P} \mapsto T_\rho^* \mathcal{P}$, a one-to-one mapping between elements of the tangent space and those of the cotangent space. The inner product $g_\rho$ is defined by the inverse of the metric tensor, often denoted as Onsager operators \citep{onsager1931reciprocal}
\begin{align}
    g_\rho(f_1, f_2) = \int f_1 G(\rho) f_2\dx{\mbtheta} = \int \phi_1 G(\rho)^{-1} \phi_2 \dx{\mbx}, \quad \phi_1 = G(\rho)f_1, \: \phi_2 = G(\rho)f_2.
\end{align}
The Wasserstein Onsager operator takes the form $G(\rho)^{-1}: \phi \mapsto -\nabla\cdot\left(\rho\nabla\phi\right)$. In the context of Wasserstein gradient flow, we define $\nabla_W E(\rho)$ as $G(\rho)^{-1} \frac{\delta E(\rho)}{\delta\rho}$, leading to the formulation:
\begin{align}
    0 = \dot{\rho}_t - G(\rho)^{-1}\frac{\delta E(\rho_t)}{\delta\rho_t} = \dot{\rho}_t + \nabla\cdot\left(\rho_t \nabla \frac{\delta E(\rho_t)}{\delta\rho_t}\right). 
\end{align}
The first variation of $\KL{\rho}{\pi}$ takes the form of $\frac{\delta\KL{\rho}{\pi}}{\delta\rho} = \log\rho/\pi + 1$, leading to the conclusion by \citet{jordan1998the} that the time evolution of the Langevin diffusion follows a curve of steepest descent in the Wasserstein space, laying the foundation for approximation using particle interaction.\par

\subsection{\gls{SVGD} as gradient flow}
\gls{PARVI} transports a set $N$ of interacting particles $\left\{\mbtheta^{(i)}_t\right\}_{1\leq i\leq N}$ towards the target distribution over time. Take \gls{SVGD} for example, denoting the empirical measure at time $t$ as $\hat{\rho}_t = \frac{1}{N} \sum_{i=1}^N \delta_{\mbtheta^{(i)}_t}$, \gls{SVGD} \citep{liu2016stein} follows the update rule 
\begin{align}
    \dot{\mbtheta}_t &= \mathbb{E}_{\mbtheta' \sim \hat{\rho}_t} \Big[ k(\mbtheta_t, \mbtheta') \nabla\log\pi(\mbtheta') + \nabla_2 k(\mbtheta_t, \mbtheta') \Big] = \mbv_\mathcal{H}(\mbtheta_t \vert\hat{\rho}_t),
\end{align}
where $k(\cdot, \cdot)$ defines a positive-definite kernel, and $\nabla_2$ denotes the gradient of the second argument. The \gls{FPE} of \gls{SVGD} can be written as
\begin{align}
    \dot{\rho}_t - \nabla\cdot\left(\rho_t \mathcal{K}_\rho \nabla\frac{\delta \KL{\rho_t}{\pi}}{\delta\rho_t}\right) = 0, \label{eq:fpe_svgd_1}
\end{align}
where $\mathcal{K}_\rho$ is an integral operator: $\mathcal{K}_\rho \mbf(\mbtheta) = \int k(\mbtheta, \mbtheta') \mbf(\mbtheta') \dx{\rho(\mbtheta)}$. We shall denote the \gls{RKHS} defined by $k$ as $\mathcal{H}$. While significantly different from \gls{LD} at first glance, \gls{SVGD} approximates the \gls{WGF} of \gls{LD} by 
\begin{itemize}
    \item kernelizing the Wasserstein Onsager operator to give $G_{\mathcal{H}}(\rho)^{-1}: \phi\mapsto -\nabla\cdot\left(\rho\mathcal{K}_\rho\nabla\phi\right)$, thus forming the Stein-Wasserstein metric $W_{\mathcal{H}}$ \citep{liu2017stein, duncan2019on};
    \item projecting the gradient flow vector field $\mbv(\rho) = -\nabla\frac{\delta \KL{\rho}{\pi}}{\delta\rho} = \nabla\log\pi - \nabla\log\rho$ onto $\mathcal{H}$ \citep{liu2019understanding}.
\end{itemize}
The interpretation of \gls{SVGD} yields valuable insights. While gradient flow \eqref{eq:fpe_svgd_1} on $(\mathcal{P}(\Omega), W_2)$ yields no closed-form energy functional \citep{chen2018a}, it behooves to absorb the operator $\mathcal{K}_\rho$ into the definition of the Stein-Wasserstein metric to guarantee a tractable energy functional. In the meantime, the kernelization trick transforms a gradient flow \eqref{eq:fpe_ld_1} simulated by a diffusion process into an approximate deterministic transportation of particles -- in other words, de-randomizes it. 
\subsection{\gls{MCMC} dynamics as fiber-gradient Hamiltonian flow}
The concept of \emph{flow} on $(\mathcal{P}(\Omega), W_2)$ is a generalization of the gradient flow: given a vector field $\mbv: \mathcal{P}(\Omega) \mapsto \bigcup_{\rho\in\mathcal{P}} T_\rho$, the flow of $\mbv$ is defined as $\dot{\rho}_t = \mbv(\rho_t)$. \citet{liu2019mcmc} interpret ``regular'' \gls{MCMC} dynamics on $(\mathcal{P}(\Omega), W_2)$ as a flow, by combining gradient flow and Hamiltonian flow on the Wasserstein space, yielding a fiber-Riemannian manifold structure of $\mathcal{P}(\Omega)$, a fiber bundle consisting of Riemannian manifolds. 
In the context of a ``regular'' \gls{MCMC} dynamics taking the form of underdamped Langevin dynamics \citep{chen2014stochastic}, the diffusion matrix $\mbA$ determines the gradient flow on each fiber, and the curl matrix $\mbC$ determines the Hamiltonian flow on the fiber bundle. The Hamiltonian flow keeps $\KL{\rho}{\pi}$ constant while encouraging fast exploration of the probability space; the fiber gradient flow minimizes the KL. The fiber-gradient Hamiltonian flow determines a \gls{PARVI} with the particle update
\begin{align}
    \dot{\mbtheta}_t = \left(\mbA(\mbtheta_t) + \mbC(\mbtheta_t)\right)\left(\nabla\log\pi(\mbtheta_t ) - \widehat{\nabla}\log\rho_t(\mbtheta_t)\right) = \mbv^{\mbA,\mbC}(\rho_t),
\end{align}
where the intractable $\widehat{\nabla}\log\rho_t$ is approximated via the Blob method \citep{carrillo2019a}. 
\subsection{Stein's method and other relevant works}
Our work enriches the practical application of Stein's method \citep{stein1972a}, which studies the class of operators that maps functions to ones with expectation zero w.r.t. a distribution $\pi$. \citet{gorham2019measuring} draw from the findings of the generator method \citep{barbour1988stein, barbour1990stein, gotze1991on} and note the same property with infinitesimal generators of Feller processes and their stationary measures, which further extends into a mapping between diffusion-based \gls{MCMC} sampling and Stein operators, noted as \emph{diffusion Stein operators}. In this work, we extend beyond sample quality measurement \citep{gorham2019measuring} and parameter estimation \citep{barp2019minimum} and establish an application of the diffusion Stein operator as a deterministic alternative of MCMC dynamics. \par
Myriad works \citep{chen2018a, liu2019understanding, zhang2020stochastic} seek alternatives to approximating the gradient flow \eqref{eq:fpe_ld_1} through kernelization. Notably, \citet{chen2018a, liu2019understanding} construct \gls{PARVI} algorithms by approximating the term $\nabla\log\rho_t(\mbtheta_t)$ in the gradient flow, namely with Blob method \citep{carrillo2019a}, kernel density estimation \citep{liu2019understanding} and Stein gradient estimator \citep{li2017gradient}. The de-randomization of underdamped \gls{LD} connects to the viewpoint of accelerating \gls{PARVI} methods. \citet{ma2019is} demonstrate that underdamped \gls{LD} \citep{chen2014stochastic} accelerates the steepest descent steps taken by the overdamped \gls{LD}, forming an analog of Nesterov acceleration for MCMC methods. \citet{wang2019accelerated} present a framework for Nesterov's accelerated gradient method in the Wasserstein space, which consists of augmenting the energy functional with the kinetic energy of an additional momentum variable. \par

\section{\gls{PARVI} for \gls{MCMC} dynamics -- a general recipe}
In this section, we outline the main contribution of this work, which traces the roadmap delineated by previous works to explore the flow interpretation, as well as the approximation by interacting particle systems of general form \gls{MCMC} dynamics \citep{ma2015a} in the form of \ito\ diffusion:
\begin{align}
    \mbf(\mbtheta) &= \frac{1}{\pi(\mbtheta)}\nabla\cdot \left[\pi(\mbtheta)\left(\mbA(\mbtheta) + \mbC(\mbtheta)\right)\right], \label{eq:sghmc1}\\
    \mbSigma(\mbtheta) &= \mbA(\mbtheta), \label{eq:sghmc2}
\end{align}
where $\mbA$ and $\mbC$ are positive-semidefinite and skew-symmetric matrix-valued functions, respectively. This general framework covers \emph{all} continuous \ito\ diffusion processes with stationary distribution $\pi$, most notably \gls{LD}, stochastic gradient \gls{HMC} \citep{chen2014stochastic}, \gls{SGNHT} \citep{ding2014bayesian} and \gls{SGRHMC} \citep{ma2015a}. \par
We demonstrate in this section that applying the fiber-gradient Hamiltonian flow structure of \gls{MCMC} dynamics to the Stein-Wasserstein metric yields a flow on $(\mathcal{P}(\Omega), W_{\mathcal{H}})$ that discretizes into \gls{GSVGD}, a \gls{PARVI} that updates particles with the diffusion Stein operator \citep{gorham2019measuring}, suggesting that the infinitesimal generator of \gls{MCMC} diffusion processes offers a ``kernel smoothing'' de-randomization. Apart from the Stein-Wasserstein metric, the analog of \gls{GSVGD} generalizing \gls{SVGD} extends to other interpretations of \gls{SVGD}, namely that \gls{GSVGD} takes steepest descent minimizing the KL-divergence in incremental transformation of particles, and that that \gls{GSVGD} projects the fiber-gradient Hamiltonian flow onto an \gls{RKHS}.\par

\begin{figure}[t]
    \centering
    \includegraphics[width=\textwidth]{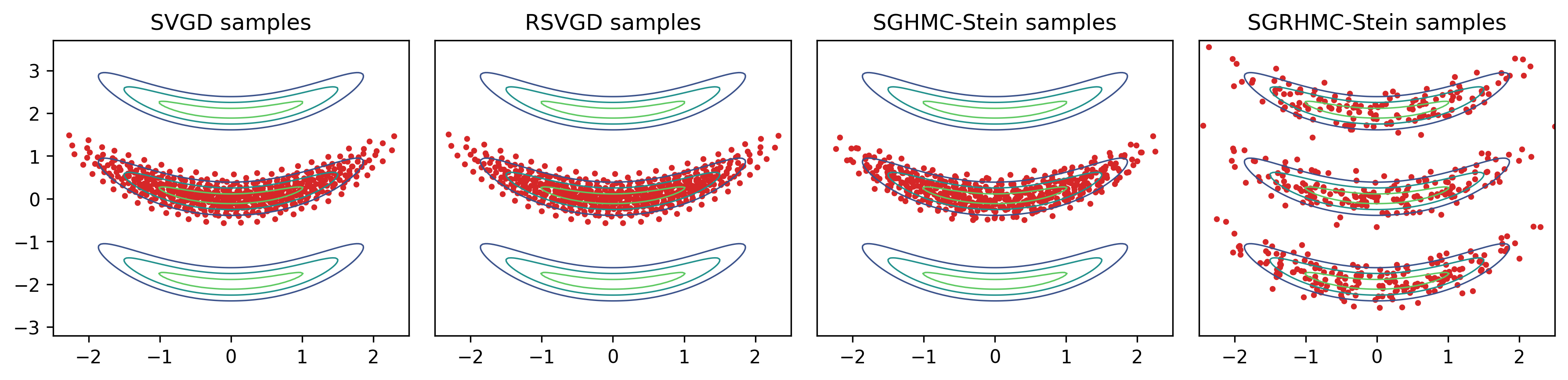}
    \caption{Particle samples of an underlying 2-dimensional correlated mixture $\pi$ extracted by SVGD \citep{liu2016stein}, Riemannian SVGD \citep{liu2017riemannian}, SGHMC-Stein and SGRHMC-Stein. While all methods explore the mode nearest to initialization, only RHMC explores all three modes.}
    \label{fig:samples}
\end{figure}

\begin{figure}[t]
    \centering
    \includegraphics[width=\textwidth]{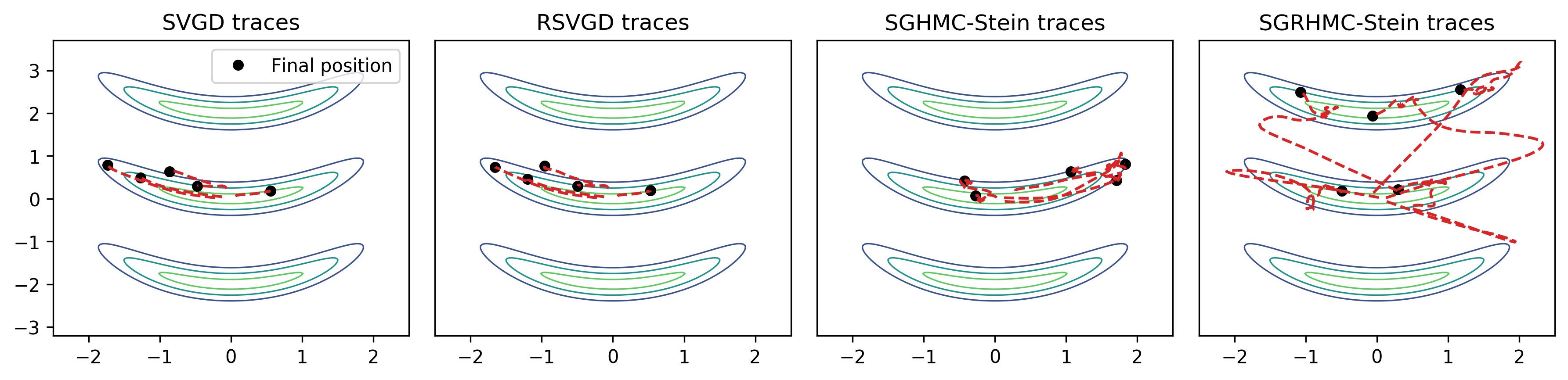}
    \caption{A visualisation of sampler trajectories on the tri-Gaussian density $\pi$. With the help of momentum variable and efficient Riemannian parametrization of $(\mbA,\mbC)$, SGRHMC-Stein aggressively explores the space, while other methods cannot escape the local component.}
    \label{fig:traces}
\end{figure}
\subsection{Fiber-gradient Hamiltonian flow on $(\mathcal{P}(\Omega), W_{\mathcal{H}})$}
Similar to \eqref{eq:fpe_ld_1}, we derive the \gls{FPE} of MCMC dynamics \eqref{eq:sghmc1}-\eqref{eq:sghmc2}
\begingroup
\allowdisplaybreaks
\begin{align}
    \dot{\rho}_t &= -\nabla\cdot \left(\rho_t\mbf \right) + (\nabla\nabla) : (\rho_t\mbA) \\
    &= -\nabla\cdot \left(\rho_t\mbf \right) + (\nabla\nabla) : (\rho_t\mbA) + \underbrace{(\nabla\nabla) : (\rho_t\mbC)}_{=0} \\
    &= \nabla\cdot\left(\rho_t \left(\mbA+\mbC\right)\nabla \log \frac{\rho_t}{\pi}\right) = \nabla\cdot\left(\rho_t \left(\mbA+\mbC\right)\nabla \frac{\delta\KL{\rho_t}{\pi}}{\delta\rho_t}\right), \label{eq:fpe_mcmc_1}
\end{align}
\endgroup
which is a curve in $\mathcal{P}(\Omega)$ following the continuity equation $\dot{\rho}_t + \nabla\cdot(\rho_t\mbv_t) = 0, \mbv_t=\left(\mbA+\mbC\right)\nabla \log \rho_t/\pi$. \par
We can derive \gls{GSVGD} from the fiber-Riemannian manifold perspective of $(\mathcal{P}(\Omega), W_{\mathcal{H}})$ \citep{liu2019mcmc}. The Riemannian structure of the Stein-Wasserstein metric induces the tangent space $T_{\rho}\mathcal{P}(\Omega) = \overline{\left\{\left.\mathcal{K}_\rho \nabla f\right\vert f\in C^{\infty}(\Omega)\right\}}^{\mathcal{H}^D}$. An orthogonal projection $\mathfrak{p}_\rho$ from $\mathcal{L}_\rho^2(\Omega)$ to $T_{\rho}\mathcal{P}$ is uniquely defined as such vector in $T_{\rho}\mathcal{P}$ that satisfies $\nabla\cdot\left(\rho \mathcal{K}_\rho\phi\right) = \nabla\cdot\left(\rho \mathfrak{p}_\rho (\phi)\right)$. Using Theorem 5 from \citet{liu2019mcmc}, we arrive at a fiber-gradient Hamiltonian flow 
\begin{align}
    \mathcal{W}_{\text{KL}_{\pi}}(\rho) &= \mathfrak{p}_\rho \left((\mbA+\mbC)\nabla \log\pi/\rho\right).
\end{align}
It is straightforward to verify that $\mbv_{\mathcal{H}}^{\mbA,\mbC}(\rho) = \mathcal{K}_\rho (\mbA+\mbC)\nabla \log\pi/\rho$ induces the same evolution of distribution as $\mathcal{W}_{\text{KL}_{\pi}}(\rho)$.
\subsection{fRH flow induces the diffusion Stein operator}
The Wasserstein flow on $(\mathcal{P}(\Omega), W_{\mathcal{H}})$ following $\dot{\rho}_t = \nabla\cdot\left(\rho_t\mbv_{\mathcal{H}}^{\mbA,\mbC}(\rho_t)\right)$ requires spatial and temporal discretization suitable for a particle update. While $\mbv_{\mathcal{H}}^{\mbA,\mbC}$ constitutes a vector field on $\mathcal{P}(\Omega)$, its calculation does not involve an explicit term of $\nabla\log\rho$ inside the expectation. 
\begingroup
\allowdisplaybreaks
\begin{align}
    \mbv_{\mathcal{H}}^{\mbA,\mbC}(\rho) &= \mathcal{K}_{\rho}\left[(\mbA+\mbC)\nabla \log\pi / \rho\right]\\
    &= \mathbb{E}_{\mbtheta'\sim\rho} k(\cdot,\mbtheta')(\mbA(\mbtheta')+\mbC(\mbtheta'))\nabla \log\pi(\mbtheta') - \int k(\cdot,\mbtheta') (\mbA(\mbtheta')+\mbC(\mbtheta'))\nabla\rho (\mbtheta')\dx\mbtheta' \notag\\
    &= \mathbb{E}_{\mbtheta'\sim\rho} (\mbA(\mbtheta')+\mbC(\mbtheta'))\nabla \log\pi(\mbtheta')k(\cdot,\mbtheta') + \underbrace{\int \nabla\cdot\left( (\mbA(\mbtheta')+\mbC(\mbtheta'))k(\cdot,\mbtheta')\right) \dx{\rho}}_{\text{integration by parts}} \notag\\
    &= \mathbb{E}_{\mbtheta'\sim\rho} \big[\underbrace{\mbf(\mbtheta')k(\cdot, \mbtheta')}_{\text{weighted drift coefficient \eqref{eq:sghmc1}}} + \underbrace{(\mbA+\mbC)(\mbtheta')\nabla_2 k(\cdot, \mbtheta')}_{\text{repulsive force}}\big]. \label{eq:gsvgd}\\
    &= \mathbb{E}_{\mbtheta'\sim\rho} \underbrace{\frac{1}{\pi} \nabla\cdot \left(\pi(\mbA+\mbC) k(\cdot, \mbtheta')\right)}_{\text{diffusion Stein operator } \mathcal{T}^{\mbA,\mbC}_\pi k(\cdot, \mbtheta')}, \label{eq:gsvgd_1}
\end{align}
\endgroup
where $\nabla_2$ refers to the gradient w.r.t. the second argument of the kernel function. Interestingly, the kernelized flow $\mbv_\mathcal{H}^{\mbA,\mbC}$ coincides with calculating the expectation of the diffusion Stein operator $\mathcal{T}^{\mbA,\mbC}_\pi$ \citep{gorham2019measuring} applied to the kernel function $k(\mbtheta,\cdot)$, a family of operators mapping functions to zero-expectation functions under an (un-normalized) distribution $\pi$. 
When $\rho_t$ is approximated by the empirical measure of its particles $\hat{\rho}_t = 1/N \sum_{i=1}^N \delta_{\mbtheta_t^{(i)}}$, we can derive the particle update of 
\begin{align}
    \dot{\mbtheta}_t &= \mbv_\mathcal{H}^{\mbA,\mbC}(\hat{\rho}_t) = \mathbb{E}_{\mbtheta'\sim\hat{\rho}_t} \mathcal{T}^{\mbA,\mbC}_\pi k(\mbtheta_t, \mbtheta') = \frac{1}{N} \sum_{i=1}^N \mathcal{T}^{\mbA,\mbC}_\pi k(\mbtheta_t, \mbtheta_t^{(i)}), \mbtheta_0^{(i)} \sim \rho_0.
\end{align}
Similar to \gls{SVGD}, the particle update of \gls{GSVGD} leverages between a weighted average of the drift vector and a repulsive force, which drives particles towards the target distribution while preventing a mode collapse. The fiber-gradient Hamiltonian flow on $(\mathcal{P}(\Omega), W_\mathcal{H})$ informs a duality between diffusion Stein operator and \gls{MCMC} dynamics, extending the use of Stein operators beyond a generalization of kernel Stein discrepancy \citep{gorham2019measuring}. 
\subsection{How does \gls{GSVGD} approximate MCMC dynamics?}
While the derivation of \gls{GSVGD} originates from the flow interpretation on $W_{\mathcal{H}}$, the analogy between \gls{SVGD} and \gls{GSVGD} expresses in equivalent alternative forms discussed by previous literature, further shedding light on the interpretation of \gls{GSVGD}. \par
Originally, the \gls{SVGD} update is expressed in the form of the functional gradient \citep{liu2016stein}. For \gls{MCMC} dynamics, we similarly derive the functional gradient with respect to the push-forward measure $\rho_\epsilon = \left(\text{id} + (\mbA+\mbC)^\top\mbv\right)_\# \rho, \mbv \in \mathcal{H}^D$, 
\begin{align}
    \nabla_{\mbv}\left.\KL{\rho_\epsilon}{\pi}\right\vert_{\mbv=\mathbf{0}} &= -\mathbb{E}_{\mbtheta'\sim\rho} \mathcal{T}^{\mbA,\mbC}_\pi k(\cdot, \mbtheta') = -\mbv_{\mathcal{H}}^{\mbA,\mbC}(\rho).
\end{align}
This presents a generalization of the discussion in \citet{liu2016stein} that \gls{GSVGD} takes incremental transformation of particles in the direction that minimizes KL-divergence, with such transformation warped by $(\mbA,\mbC)$ and constrained by \gls{RKHS}. \par
Alternatively, \citet{liu2019understanding} explains \gls{SVGD} as the projection of the original gradient flow direction $\mbv(\rho_t) = \nabla\log\pi/\rho_t$ onto an \gls{RKHS}. We similarly generalize this explanation for \gls{SVGD}, as it projects $\mbv^{\mbA,\mbC}(\rho_t) = (\mbA+\mbC)\nabla\log\pi/\rho_t \in \mathcal{L}_\rho^2$ onto $\mathcal{H}^D$, as 
\begin{align}
    \mbv_{\mathcal{H}}^{\mbA,\mbC}(\rho) = \maxargmax_{\mbv\in\mathcal{H}^D, \norm{\mbv}_{\mathcal{H}^D}=1}
    \langle \mbv^{\mbA,\mbC}(\rho), \mbv\rangle_{\mathcal{L}_\rho^2}.
\end{align}
The analogy of \gls{GSVGD} goes further when we consider the transportation of $N$ particles as inferring a joint distribution $\pi^{\otimes N}$, which will be covered in the supplements.
\subsection{\gls{GSVGD} in practice}
We can apply \gls{GSVGD} updates \eqref{eq:gsvgd} to de-randomize myriad \gls{MCMC} dynamics (e.g., Table \ref{tab:table1}). Notably, \gls{GSVGD} establishes the previously unexplored duality between Riemannian Langevin diffusion \citep{girolami2011riemann} and Riemannian \gls{SVGD} \citep{liu2017riemannian}.\par
To fully harness the capacity of our framework, we can introduce auxiliary (momentum) variables $\mbr$ to help with the exploration of probability space, namely to augment the target distribution as $\pi(\mbtheta,\mbr) = \pi(\mbtheta)\mathcal{N}(\mbr \vert\mathbf{0}, \mbSigma)$, therefore achieving de-randomized \gls{PARVI} variant of \gls{SGHMC} \citep{chen2014stochastic}. Further leveraging a positive definite $\mbG(\mbtheta)$ to efficiently explore the target distribution yields a \gls{PARVI} variant of \gls{SGRHMC} \citep{ma2015a}.\par
When \gls{GSVGD} is used in conjunction with auxiliary momentum variables, we can employ symmetric splitting for leapfrog-like steps for \gls{GSVGD}s de-randomizing with momentum, 
\begin{align}
    \mbr_{k+1/2}^{(i)} &= \mbr_k^{(i)} + \frac{\epsilon}{2}\mbv_{\mathcal{H}}^{\mbA,\mbC}\left(\left.\mbr_k^{(i)}\right\vert\hat{\rho}(\mbtheta_k, \mbr_k)\right),\\
    \mbtheta_{k+1}^{(i)} &= \mbtheta_k^{(i)} + \epsilon \mbv_{\mathcal{H}}^{\mbA,\mbC}\left(\left.\mbtheta_k^{(i)}\right\vert\hat{\rho}(\mbtheta_k, \mbr_{k+1/2})\right),\\
    \mbr_{k+1}^{(i)} &= \mbr_{k+1/2}^{(i)} + \frac{\epsilon}{2}\mbv_{\mathcal{H}}^{\mbA,\mbC}\left(\left.\mbr_{k+1/2}^{(i)}\right\vert\hat{\rho}(\mbtheta_{k+1}, \mbr_{k+1/2})\right).
\end{align}
\section{Experiments}
\begin{figure}[t]
    \centering
    \includegraphics[width=\textwidth]{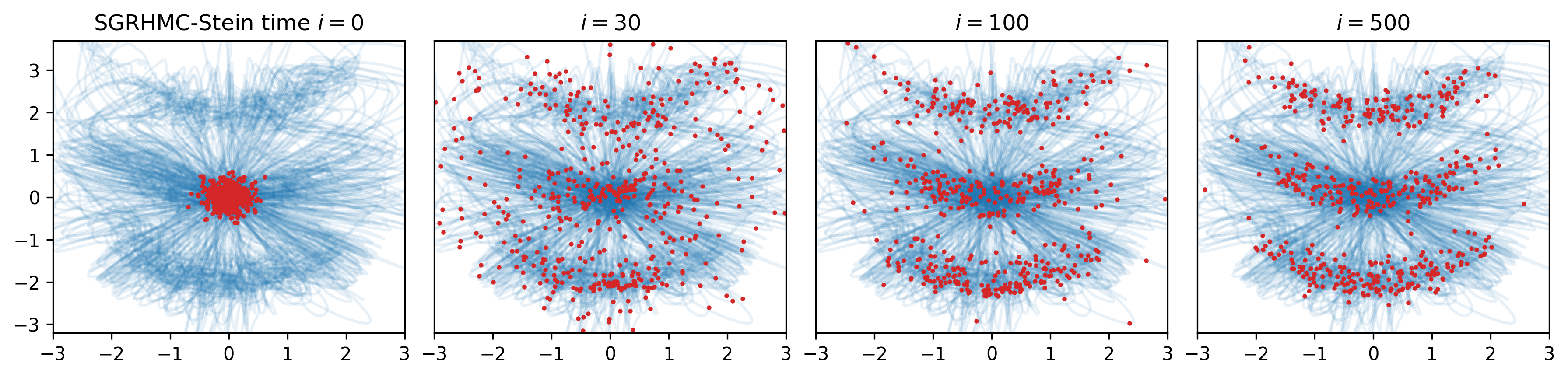}
    \caption{Evolution of RHMC dynamics (\textcolor{blue}{blue lines}) over four highlighted time increments $k$ (\textcolor{red}{red points}) on tri-Gaussian density. The RHMC explores the space by an aggressive outward expansion followed by convergence to the tri-Gaussian density $\pi$.}
    \label{fig:evolution}
\end{figure}
\subsection{Stein \gls{PARVI} on toy data}
To demonstrate the efficacy of our methods, we use various \gls{PARVI} to infer a 3-component mixture of crescent-shaped target measure inspired by \citet{ma2015a}. The parametrizations of Riemannian \gls{SVGD} and SGRHMC-Stein coincide with those in  \citet{ma2015a}. While all methods explore the nearest mode well (See figure \ref{fig:samples}), only the \gls{SGRHMC}-Stein manage to effectively use the Riemannian formulation of $\left(\mbA, \mbC\right)$ to explore the 2 other modes (see figure \ref{fig:traces}), showcasing the gain in efficiency with application-specific dynamics.
\begin{table}[b]
    \centering
    \resizebox{\textwidth}{!}{
    \begin{tabular}{lcccccccc}
\toprule
Method &         boston &       concrete &         energy &        kin8nm &          power &          yacht &           year &        protein \\
\midrule
LD          &  -2.52 (0.15) &  -3.16 (0.04) &  -2.36 (0.06) &  0.10 (0.03) &  -2.85 (0.03) &  -1.60 (0.08) &  -3.69 (0.01) &  -3.05 (0.01) \\
SVGD        &  -2.60 (0.46) &  -3.11 (0.10) &  -1.95 (0.07) &  1.03 (0.25) &  -2.82 (0.03) &  -1.66 (0.17) &  -3.61 (0.00) &  -2.91 (0.01) \\
Blob        &  -2.52 (0.32) &  -3.19 (0.07) &  -1.67 (0.03) &  0.95 (0.03) &  -2.81 (0.04) &  -1.34 (0.07) &  -3.60 (0.00) &  -2.93 (0.01) \\
SGHMC-Blob    &  \textbf{-2.42 (0.12)} &  \textbf{-2.95 (0.01)} &  -1.36 (0.26) &  1.23 (0.02) &  -2.77 (0.04) &  \textbf{-0.73 (0.46)} &  -3.60 (0.02) &  -3.73 (0.16) \\
SGNHT       &  -2.59 (0.13) &  -3.41 (0.05) &  -2.44 (0.04) &  0.75 (0.02) &  -2.86 (0.02) &  -2.64 (0.04) &  -3.69 (0.00) &  -3.01 (0.01) \\
DE          &  -2.68 (0.63) &  -2.96 (0.15) &  \textbf{-0.45 (0.14)} &  1.16 (0.02) &  -2.80 (0.03) &  -1.04 (0.40) &  -3.68 (0.00) &  -3.00 (0.01) \\
SGHMC-Stein &  -2.81 (0.71) &  -3.04 (0.25) &  -1.40 (0.82) &  \textbf{1.25 (0.02)} &  \textbf{-2.76 (0.04)} &  -0.86 (0.33) &  \textbf{-3.59 (0.00)} &  -2.90 (0.01) \\
SGNHT-Stein &  -2.49 (0.30) &  -2.97 (0.11) &  \textbf{-0.44 (0.10)} &  1.24 (0.03) &  -2.78 (0.04) &  -0.85 (0.21) &  \textbf{-3.59 (0.00)} &  \textbf{-2.85 (0.01)} \\
\bottomrule
\end{tabular}
}
    \caption{The test log-likelihood (higher is better) results for selected datasets in UCI repository. The results are reported in mean (standard deviation) form averaged over 20 runs in the first 6 columns and 6 runs in the last 2, with the best performing model marked in boldface.}
    \label{tab:uci_results}
\end{table}
\subsection{\gls{GSVGD} on Bayesian neural networks}
\begin{figure}
    \centering
    \includegraphics[width=0.8\textwidth]{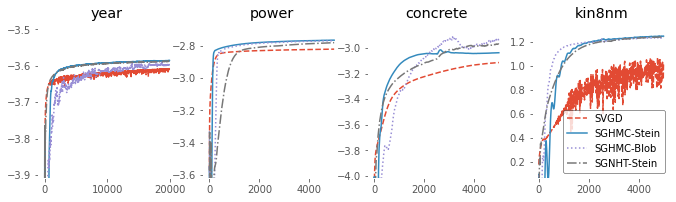}
    \caption{The trace plots of test log-likelihoods of BNN experiments: the x-axis determines the iterations. The figures show the effect of the momentum variables in SGHMC-Stein and SGNHT-Stein in exploring the posterior space, and quickly converging to better optima. The jaggedness of \gls{SVGD} showcases that while \gls{SVGD} empirically works with stochastic gradient, it is less robust with stochastic gradient noise.}
    \label{fig:bnn_trace}
\end{figure}

We apply \gls{GSVGD} of advanced \gls{MCMC} dynamics to the inference of Bayesian neural networks, taking a simple structure of one hidden layer, and an output with Gaussian likelihood. We opt for the fully Bayesian specification of BNN, where the precision parameter of the weight prior and the Gaussian likelihood follows $\text{Gamma}(1, 0.1)$. The \gls{GSVGD} variant used in the experiments include \gls{SGHMC} \citep{chen2014stochastic} and \gls{SGNHT} \citep{ding2014bayesian}, with target measure $\pi(\mbtheta, \mbr, \mbxi) = \pi(\mbtheta)\mathcal{N}(\mbr\vert \mathbf{0}, \sigma^2\mbI)\mathcal{N}(\mbxi\vert A\mathbf{1}, \mu^{-1}\mbI)$, and $\mbA = \left(\begin{matrix}\mathbf{0}&\mathbf{0}&\mathbf{0}\\ \mathbf{0}& A\mbI& \mathbf{0}\\
\mathbf{0}&\mathbf{0}&\mathbf{0}\end{matrix}\right), \mbC = \left(\begin{matrix}\mathbf{0} & -\mbI & \mathbf{0}\\  \mbI & \mathbf{0} & (\mu \sigma^2)^{-1}\text{diag}(\mbr)\\ \mathbf{0} & -(\mu \sigma^2)^{-1}\text{diag}(\mbr) & \mathbf{0}\end{matrix}\right)$. As a \gls{MCMC} form suitable when used in conjunction with stochastic gradients, \gls{SGNHT} takes additional temperature parameter $\mbxi$. The log-likelihood results in Table \ref{tab:uci_results} demonstrates that the de-randomized \gls{MCMC} dynamics achieve superior predictive performance than their \gls{LD} counterparts, largely thanks to the exploration of probability space from the momentum variables (See figure \ref{fig:bnn_trace}), with a notable gain from symmetric splitting. \gls{SGNHT}-Stein methods show robustness to hyperparameter selection and stochastic gradient noise. 
\section{Discussion}
In this section, we discuss open questions pertinent to the ``optimality'' of the diffusion Stein operator as a \gls{PARVI} update, and a direction for future work regarding the fiber-gradient Hamiltonian flow of the chi-squared divergence, inspired by \citet{chewi2020svgd}.
\subsection{Alternatives to the diffusion Stein operator}
The vector field $\mbv^{\mbA,\mbC}(\rho) = \left(\mbA+\mbC\right)\nabla\log\frac{\pi}{\rho}$ is not the unique vector field that induces the time evolution of \gls{MCMC} dynamics: indeed, two vector fields $\mbv_1, \mbv_2$ induce the same curve in $\mathcal{P}(\Omega)$ when they differ by a divergence-free field: $\nabla\cdot\left(\rho (\mbv_1(\rho)-\mbv_2(\rho))\right) = 0$. Notably, \citet{liu2019mcmc} propose an alternative vector field $\tilde{\mbv}^{\mbA,\mbC}(\rho) = \mbv^{\mbA,\mbC}(\rho) - \mbC\nabla\log\rho + \nabla\cdot \mbC$ \footnote{In fact, a divergence-free vector field $\mbC_0\nabla\log\rho - \nabla\cdot \mbC_0$ can be constructed out of an arbitrary skew-symmetric matrix-valued function $\mbC_0$}. A projection onto \gls{RKHS} yields a Stein \gls{PARVI} with update
\begin{align}
    \dot{\mbtheta}_t &= \mathbb{E}_{\rho_t}\left[ \frac{1}{\pi} \nabla\cdot \left[\pi\left(\mbA+\mbC\right)\right] k(\mbtheta_t, \cdot)+ \mbA \nabla_2 k(\mbtheta_t, \cdot)\right], \label{eq:gsvgd_alt}
\end{align}
consistent with the standard formulation infinitesimal generator. However, the expectation in \eqref{eq:gsvgd_alt} notably does not converge to zero when $\rho_t = \pi$ \citep{gorham2019measuring}, suggesting that the particles will keep rotating along the trajectories of a divergence-free vector field, as opposed to stopping, in the equilibrium state. Interested readers can find experiments with respect to this alternative form in the supplementary materials.
\subsection{Generalizing LAWGD with the diffusion Stein operator}
The application of the diffusion Stein operator generalizes \gls{LAWGD} \citep{chewi2020svgd}, which views \gls{SVGD} as a kernelized Wasserstein gradient flow driven by the chi-squared divergence. Given the first variation of the chi-squared divergence $\frac{\delta\chisq{\rho}{\pi}}{\delta\rho} = \frac{2{\rho}}{{\pi}}$, we can verify that the vector field $\mbv_{\mathcal{H}}^{\mbA,\mbC}=\mathcal{K}_\rho\left[\left(\mbA+\mbC\right)\nabla\log\frac{{\pi}}{{\rho}}\right]$ is rewritten as 
\begin{align}
    2\mbv_{\mathcal{H}}^{\mbA,\mbC} &= \mathcal{K}_\pi\left[\left(\mbA+\mbC\right)\nabla\frac{2{\pi}}{{\rho}}\right] = \mathcal{K}_{\pi} \left[(\mbA+\mbC) \nabla \frac{\delta\chisq{\rho_t}{\pi}}{\delta\rho_t}\right].
\end{align}
The chi-squared divergence perspectives enriches \gls{GSVGD} as a kernelized fiber-gradient Hamiltonian flow minimizing the chi-squared divergence. Replacing the vector field $\mathcal{K}_{\pi} \left[(\mbA+\mbC) \nabla \frac{\dx{\rho}}{\dx{\pi}}\right]$ with $\mbu=\nabla\cdot\left(\mathcal{K}_{\pi} \left[(\mbA+\mbC) \frac{\dx{\rho}}{\dx{\pi}}\right]\right)$, we get the dissipation formula for the KL-divergence 
\begin{align}
    \frac{\dx{\KL{\rho_t}{\pi}}}{\dx{t}} &= -\mathbb{E}_\pi \left\langle\nabla\frac{\rho}{\pi}, \nabla\cdot\left(\mathcal{K}_{\pi} \left[(\mbA+\mbC) \frac{\rho}{\pi}\right]\right) \right\rangle= -\mathbb{E}_\pi \left[\frac{\rho}{\pi} \mathcal{A}^{\mbA,\mbC}_\pi \mathcal{K}_\pi \frac{\rho}{\pi}\right],
\end{align}
where $\mathcal{A}^{\mbA,\mbC}_\pi$ represents the infinitesimal generator that induces the diffusion Stein operator: $\mathcal{A}^{\mbA,\mbC}_\pi g = \frac{1}{2\pi}\nabla\cdot\left(\pi(\mbA+\mbC)\nabla g\right)$. Following the construction of \gls{LAWGD}, we choose the kernel so that $\mathcal{K}_\pi = \big(\mathcal{A}^{\mbA,\mbC}_\pi\big)^{-1}$. While it remains to be seen what kernel induces this equality, the diffusion Stein operator yields a more flexible formulation, while retaining the superior convergence of \gls{LAWGD}.

\section{Conclusion}
We further the application of the Stein operator and its generalizations by proposing \gls{GSVGD}, an interacting particle system transporting a particle set towards a target distribution $\pi$ in an emulation of the corresponding \gls{MCMC} dynamics. We showcase theoretically and empirically that \gls{GSVGD} helps augment the well-researched field of efficient \gls{MCMC} dynamics with deterministic interacting particle systems with high-quality samples.


\bibliography{refs}
\bibliographystyle{unsrtnat}

\end{document}


\maketitle
In the supplementary material, we expand on the background details and derivations mentioned in the paper. 
\section{Additional derivations for the Wasserstein gradient flow}
\subsection {Stochastic differential equations and its Fokker-Planck equation}
The general SDE
\begin{align}
    \dx{\mbtheta_t} = \underbrace{\mbf(\mbtheta_t)}_{\text{drift}}\dx{t} + \underbrace{\sqrt{2\mbSigma(\mbtheta_t)}}_{\text{diffusion}} \underbrace{\dx{\mbW_t}}_{\text{Wiener process}}
\end{align}
with an initial distribution $\mbtheta_0 \sim \rho_0$ defines the evolution of a random variable $\bt_t \in \mathbb{R}^D$ over time $t \in \mathbb{R}_+$. The evolution of the marginal distribution $\rho_t$ is given by the Fokker-Planck equation
\begin{align}
    \dot{\rho}_t(\mbtheta) &= - \sum_{i=1}^D \frac{\partial}{\partial \theta_i} \rho_t(\bt) f_i(\bt) + \sum_{i,j=1}^{D} \frac{\partial^2 }{\partial \theta_i \partial \theta_j} \rho_t(\bt) \Sigma_{ij}(\bt) \\
    &= - \nabla \cdot (\rho_t \f) + (\nabla^\top \nabla) : (\rho_t \bS),
\end{align}
where we use the dyadic vector notation
\begin{align}
    \mathbf{A} : \mathbf{B} &= \mathrm{tr} \{ \mathbf{A}^\top \mathbf{B} \}.
\end{align}
The double-dot notation indicates a sum over all element-wise products, or the sum of all second order partial derivatives. We can verify this with 
\begin{align}
    (\nabla^\top \nabla) : (\rho_t \bS) &= \mathrm{tr} \{ (\nabla^\top \nabla)^\top (\rho_t \bS) \} \\
    &= \mathrm{tr} \{ (\nabla \nabla^\top) (\rho_t \bS) \} \\
    &= \langle \nabla\nabla^\top, \rho_t \bS \rangle_F \\
    &= \sum_{i,j} \frac{\partial^2}{\partial \theta_i \partial \theta_j} \rho_t(\bt_t) \bS(\bt_t).
\end{align}

\subsection{FPE of Langevin dynamics}

The Langevin dynamics has drift $\f(\bt_t) = \nabla \log \pi(\bt_t)$ and diffusion $\bS(\bt_t) = \I$. The Fokker-Planck equation for Langevin dynamics then simplifies into 
\begin{align}
    \dot{\rho}_t &= - \nabla \cdot (\rho_t \f) + (\nabla\nabla^\top) : (\rho_t \I) \\
    &= - \nabla \cdot (\rho_t \nabla \log \pi) + \Delta \rho_t,
\end{align}
where the Laplacian is defined as
\begin{align}
    \Delta \rho_t &= \nabla^2 \rho_t = \sum_{i=1}^D \frac{\partial^2 \rho_t(\mbtheta)}{\partial \theta_i^2}.
\end{align}

To derive the FPE we begin by using the Laplacian identity $\Delta \rho_t = \nabla \cdot (\nabla \rho_t)$, resulting in
\begin{align}
    \dot{\rho}_t &= \Delta \rho_t - \nabla \cdot \big(\rho_t \nabla \log \pi\big) \\
    &= \nabla \cdot \big(\nabla \rho_t\big) - \nabla \cdot \big(p \nabla \log \pi\big) \\
    &= \nabla \cdot \big(\nabla \rho_t - \rho_t \nabla \log \pi\big) \\
    &= \nabla \cdot \big(\rho_t \nabla \log \rho_t - \rho_t \nabla \log \pi\big) \\
    &= \nabla \cdot \left(\rho_t \nabla \log \frac{\rho_t}{\pi}\right),
\end{align}
where we used the identity $\nabla \log \rho = \frac{\nabla \rho}{\rho}$ to expand $\nabla \rho = \rho \nabla \log \rho$. Notice that we did not need to use the common identity $\nabla \cdot (\rho \f) = \rho \nabla \cdot \f + (\nabla \rho) \cdot \f$, nor expand with $\nabla \log \pi = - \nabla H$.

We also notice that the first variation of the Kullback-Leibler divergence has the form $\frac{\delta \KL{\rho_t}{\pi}}{\delta\rho_t} = \log\rho_t/\pi+1$, so that $\nabla\frac{\delta \KL{\rho_t}{\pi}}{\delta\rho_t}=\nabla\log\rho_t-\nabla\log\pi$, yielding the final result
\begin{align}
    \dot{\rho}_t &= \nabla \cdot \left( \rho_t \nabla \frac{\delta \KL{\rho_t}{\pi}}{\delta \rho_t}\right).
\end{align}
In the original variational formulation for the Wasserstein gradient flow \citep{jordan1998the}, the authors prove the weak convergence of a discrete gradient flow taking discrete steps at interval $h$, such that each $\rho_{kh}, k=1, 2,\hdots,$ follows the minimization
\begin{align}
    \rho_{kh} = \argmin_{\rho \in \mathcal{P}(\Omega)} \quad  \KL{\rho}{\pi} + \frac{1}{2h} W_2^2(\rho, \rho_{(k-1)h}). 
\end{align}
$\left(\rho_{kh}\right)_{k=1}^\infty$ weakly converges to the Fokker-Planck equation as $h\downarrow 0$.
\subsection{The Stein-Wasserstein metric}
With the Onsager operator defined as $G(\rho)^{-1}: \phi \mapsto -\nabla\cdot\left(\rho\mathcal{K}_\rho\nabla\phi\right)$, the Stein-Wasserstein metric between $\rho_0$ and $\rho_1$ is defined using the geometric action function
\begin{align}
    W_{\mathcal{H}}^2(\rho_0, \rho_1) = \inf_{\phi,\rho_t} &\left\{\int_0^1 \int \norm{\mathcal{K}_{\rho_t}\nabla\phi_t }_{\mathcal{H}}^2\dx{t}\: : \: \dot{\rho}_t + \nabla \cdot (\rho_t \mathcal{K}_{\rho_t}\nabla\phi_t) = 0\right\},
\end{align}
where $\mathcal{K}_\rho$ is the integral operator 
\begin{align}
\mathcal{K}_\rho \mbf(\mbtheta) = \int k(\mbtheta', \mbtheta) \mbf(\mbtheta')\text{d}\rho(\mbtheta'),
\end{align}
which smoothens the function $\mbf$ over a similar parameters $\mbtheta'$ from the density $\rho$ according to the kernel $k(\mbtheta', \mbtheta)$. The Stein-Wasserstein metric follows the definition that the distance between two points in $\mathcal{P}(\Omega)$ consists of the length the shortest arc connecting the two points, parametrized by $\phi_t$. 

\section{Additional derivations for the analysis of MCMC dynamics}
To recap, MCMC dynamics consist of positive semi-definite matrix-valued function $\mbA$ and skew-symmetric matrix-valued function $\mbC$ with 
\begin{align}
    \mbf(\mbtheta) &= \frac{1}{\pi(\mbtheta)}\nabla\cdot\left(\pi(\mbtheta)(\mbA(\mbtheta)+\mbC(\mbtheta))\right),\\
    \mbSigma(\mbtheta) &= \mbA(\mbtheta)
\end{align}
with matrix properties 
\begin{align}
    \mbx^\top \mbA \mbx \ge 0, \qquad \forall \mbx \in  \mathbb{R}^D \\
    \mbC^\top = -\mbC, \quad C_{ij} = -C_{ji}, \quad \mathrm{diag} \, \mbC = \mathbf{0}.
\end{align}

\subsection{Fokker-Planck equation and its equivalent forms}

We derive the Fokker-Planck equation of SDEs with the above MCMC dynamics form as
\begin{align}
    \dot{\rho}_t(\mbtheta) &= -\sum_{i=1}^D \frac{\partial}{\partial\theta_i} \rho_t(\mbtheta) f_i(\mbtheta) + \sum_{i,j=1}^D \frac{\partial^2}{\partial\theta_i\partial\theta_j} \rho_t(\mbtheta)A_{ij}(\mbtheta)\\
    &= -\sum_{i=1}^D \frac{\partial}{\partial\theta_i} \rho_t(\mbtheta) f_i(\mbtheta) + \sum_{i,j=1}^D \frac{\partial^2}{\partial\theta_i\partial\theta_j} \rho_t(\mbtheta)A_{ij}(\mbtheta) + \underbrace{\sum_{i,j=1}^D \frac{\partial^2}{\partial\theta_i\partial\theta_j} \rho_t(\mbtheta)C_{ij}(\mbtheta)}_{=0}\\
    &= -\nabla\cdot \left(\frac{\rho_t}{\pi}\nabla\cdot\left(\pi(\mbA+\mbC)\right)\right) + (\nabla^\top\nabla):(\rho_t(\mbA+\mbC))\\
    &= -\nabla\cdot \left(\rho_t\left[(\mbA+\mbC)\nabla\log\pi(\mbtheta) + \nabla\cdot(\mbA+\mbC)\right]\right) + (\nabla^\top\nabla):(\rho_t(\mbA+\mbC))\\
    &= -\nabla\cdot \left(\rho_t\left[(\mbA+\mbC)\nabla\log\pi(\mbtheta) + \nabla\cdot(\mbA+\mbC)\right]\right) + \nabla\cdot\left(\rho_t\left[(\mbA+\mbC)\nabla\log\rho_t + \nabla\cdot(\mbA+\mbC)\right]\right)\\
    &= \nabla\cdot \left(\rho_t(\mbA+\mbC)\nabla\log\frac{\rho_t}{\pi}\right), \label{eq:mcmc_fpe}
\end{align}
where the skew-symmetric addition is justified via
\begin{align}
    \sum_{i,j=1}^D \frac{\partial^2}{\partial\theta_i\partial\theta_j} \rho_t(\mbtheta) C_{ij}(\mbtheta) &= \underbrace{\sum_{i,j=1}^D \frac{\partial^2\rho_t(\mbtheta)}{\partial\theta_i\partial\theta_j} C_{ij}(\mbtheta)}_{=0} + \underbrace{\sum_{i,j=1}^D \rho_t(\mbtheta)\frac{\partial^2 C_{ij}(\mbtheta)}{\partial\theta_i\partial\theta_j}}_{=0}\\
    &+ \sum_{i,j=1}^D \left[\frac{\partial\rho_t(\mbtheta)}{\partial\theta_i}\frac{\partial C_{ij}(\mbtheta)}{\partial\theta_j} + \frac{\partial\rho_t(\mbtheta)}{\partial\theta_j}\frac{\partial C_{ij}(\mbtheta)}{\partial\theta_i}\right]\\
    &= \sum_{i,j=1}^D \frac{\partial\rho_t(\mbtheta)}{\partial\theta_i}\frac{\partial C_{ij}(\mbtheta)}{\partial\theta_j} + \sum_{j,i=1}^D \frac{\partial\rho_t(\mbtheta)}{\partial\theta_j}\frac{\partial C_{ij}(\mbtheta)}{\partial\theta_i}\\
    &= \sum_{i,j=1}^D \frac{\partial\rho_t(\mbtheta)}{\partial\theta_i}\frac{\partial C_{ij}(\mbtheta)}{\partial\theta_j} + \sum_{j,i=1}^D \frac{\partial\rho_t(\mbtheta)}{\partial\theta_j}\frac{\partial -C_{ji}(\mbtheta)}{\partial\theta_i} = 0.
\end{align}
The final form \eqref{eq:mcmc_fpe} corresponds to a generalization of the continuity equation $\dot{\rho}_t + \nabla\cdot \left(\rho_t(\mbA+\mbC)\nabla\phi_t\right) = 0$.
\subsection{The diffusion Stein operator}
Originally, the Stein's identity \citep{stein1972a} maps sufficiently regular functions $\mbPhi: \mathbb{R}^D\mapsto \mathbb{R}^D$ to $\mathcal{T}\mbPhi(\mbtheta) = \nabla\log\pi(\mbtheta)\cdot \mbPhi(\mbtheta) + \nabla\cdot\mbPhi(\mbtheta)$. The function $\mathcal{T}\mbPhi$ has expectation zero under the target measure $\pi$: $\mathbb{E}_\pi \mathcal{T}\mbPhi = 0$, yielding the original Stein's identity. As a generalization, \citet{gorham2019measuring} discuss the application of infinitesimal generator of MCMC dynamics as a means to discover operators sharing the same property. Infinitesimal generators of Feller processes describes the perturbation of functions:
\begin{align}
    \mathcal{A}u(\mbtheta) &= \lim_{t\rightarrow 0} \frac{\mathbb{E}\left[u(\mbtheta_t)\vert\mbtheta_0=\mbtheta\right] - \mbu(\mbtheta)}{t}.
\end{align}
For MCMC dynamics with parametrization $(\mbA,\mbC)$, the infinitesimal generator is calculated as 
\begin{align}
    \left(\mathcal{A}_\pi^{\mbA,\mbC} u\right)(\mbtheta) &= \frac{1}{2\pi(\mbtheta)} \nabla\cdot\left(\pi(\mbtheta)(\mbA(\mbtheta)+\mbC(\mbtheta))\nabla u(\mbtheta)\right).
\end{align}
And the \emph{diffusion Stein operator}, denoted as $\mathcal{T}_\pi^{\mbA,\mbC}$ in this work, is defined by substituting $\frac{\nabla u}{2}$ with a vector-valued function $\mbf$. GSVGD is defined as $\mathbb{E}_\rho \mathcal{T}_\pi^{\mbA,\mbC} k(\cdot,\mbtheta)$.
\subsection{GSVGD as functional gradient in RKHS for incremental transformations}
Functional gradient for a functional $F[\cdot]$ is defined as such $\nabla_{\mbf} F[\mbf]$ that satisfies $F[\mbf + \epsilon\mbg] = F[\mbf] + \langle\nabla_\mbf F[\mbf], \mbg\rangle_{\mathcal{H}^D} + O(\epsilon^2)$. The particle update of GSVGD can be seen as the functional gradient with respect to the push-forward measure $\rho_{\epsilon} = \left(\text{id} + (\mbA+\mbC)^\top\mbf\right)_\# \rho = \left(\text{id} + (\mbA-\mbC)\mbf\right)_\# \rho$. Following the proof of Theorem 3.3 in \citet{liu2016stein}, we define $F[\mbf] = \KL{\rho_\epsilon}{\pi}=\KL{\left(\text{id} + (\mbA-\mbC)\mbf\right)_\#\rho}{\pi} = \KL{\rho}{\left(\text{id} + (\mbA-\mbC)\mbf\right)^{-1}_\#\pi}$
\begin{align}
    F[\mbf + \epsilon\mbg] &= \mathbb{E}_\rho\left[\log\rho(\mbtheta) - \log \pi (\mbtheta+(\mbA-\mbC)(\mbf+\epsilon\mbg)) - \log\det \left(\mbI + \underbrace{\nabla((\mbA-\mbC)(\mbf+\epsilon\mbg)}_{\text{Jacobian matrix}})\right)\right].
\end{align}
We then have
\begin{align}
    F[\mbf + \epsilon\mbg] - F[\mbf] &= -\underbrace{\mathbb{E}_\rho\left[\log \frac{\pi (\mbtheta+(\mbA-\mbC)(\mbf+\epsilon\mbg))}{\pi(\mbtheta+(\mbA-\mbC)\mbf)}\right]}_{\Delta_1} - \underbrace{\mathbb{E}_\rho\left[\log\frac{\det \left(\mbI + \nabla\left((\mbA-\mbC)(\mbf+\epsilon\mbg)\right)\right)}{\det \left(\mbI + \nabla((\mbA-\mbC)\mbf)\right)}\right]}_{\Delta_2}.
\end{align}
\begin{align}
    \Delta_1 &= \mathbb{E}_\rho \left[\log \pi (\mbtheta+(\mbA-\mbC)(\mbf+\epsilon\mbg)) - \log\pi(\mbtheta+(\mbA-\mbC)\mbf)\right] \\
    &= \epsilon \mathbb{E}_{\rho}\left[\nabla\log\pi(\mbtheta+(\mbA-\mbC)\mbf)\cdot(\mbA-\mbC)\mbg \right] + O(\epsilon^2)\\
    &= \mathbb{E}_{\rho}\left[(\mbA+\mbC)\nabla\log\pi(\mbtheta+(\mbA(\mbtheta)-\mbC(\mbtheta))\mbf(\mbtheta))\right]\cdot\mbg + O(\epsilon^2)\\
    &= \mathbb{E}_{\mbtheta\sim\rho}\left[(\mbA(\mbtheta)+\mbC(\mbtheta))\nabla\log\pi(\mbtheta+(\mbA(\mbtheta)-\mbC(\mbtheta))\mbf(\mbtheta))\right]\cdot\langle k(\mbtheta,\cdot), \mbg(\cdot)\rangle_{\mathcal{H}^D} + O(\epsilon^2)\\
    &= \left\langle\mathbb{E}_{\mbtheta\sim\rho}\left[(\mbA(\mbtheta)+\mbC(\mbtheta))\nabla\log\pi(\mbtheta+(\mbA(\mbtheta)-\mbC(\mbtheta))\mbf(\mbtheta))\right] k(\mbtheta,\cdot), \mbg(\cdot)\right\rangle_{\mathcal{H}^D} + O(\epsilon^2),\\
    \Delta_2 &= \mathbb{E}_\rho\left[\log\det \left(\mbI + \nabla((\mbA-\mbC)(\mbf+\epsilon\mbg)) \right) - \log \det \left(\mbI + \nabla((\mbA-\mbC)\mbf)\right)\right]\\
    &= \epsilon \mathbb{E}_\rho\left[\left(\mbI + \nabla((\mbA-\mbC)\mbf)\right)^{-1} : \nabla\left((\mbA-\mbC)\mbg\right)\right] + O(\epsilon^2)\\
    &= \epsilon\mathbb{E}_\rho\left[\left(\mbI + \nabla((\mbA-\mbC)\mbf)\right)^{-1} : \left\{\underbrace{\left(\mbA-\mbC\right)\nabla\mbg}_{\Delta_3} + \underbrace{\mbM}_{\Delta_4}\right\} \right] + O(\epsilon^2),
\end{align}
where $\mbM_{ij} = \sum_{\ell=1}^D \frac{\partial(\mbA-\mbC)_{i\ell}}{\partial\mbtheta_j}g_\ell$. We have
\begin{align}
    \mbM_{ij}(\mbtheta) &= \sum_{\ell=1}^D \frac{\partial(\mbA(\mbtheta)-\mbC(\mbtheta))_{i\ell}}{\partial\mbtheta_j}g_\ell(\mbtheta)\\
    &= \sum_{\ell=1}^D \frac{\partial(\mbA(\mbtheta)-\mbC(\mbtheta))_{i\ell}}{\partial\mbtheta_j}\langle k(\mbtheta,\cdot), g_\ell(\cdot)\rangle_{\mathcal{H}}\\
    &= \langle \mbh^{(ij)} k(\mbtheta,\cdot), \mbg(\cdot)\rangle_{\mathcal{H}^D},
\end{align}
where $\mbh^{(ij)}_\ell = \frac{\partial(\mbA(\mbtheta)-\mbC(\mbtheta))_{i\ell}}{\partial\mbtheta_j}$.
\begin{align}
    \Delta_3 &= \epsilon\mathbb{E}_\rho \left[\left(\mbI + \nabla((\mbA-\mbC)\mbf)\right)^{-1} : \left\{\left(\mbA-\mbC\right)\nabla\mbg\right\}\right]\\
    &= \epsilon\mathbb{E}_\rho \left[\left(\mbI + \nabla((\mbA-\mbC)\mbf)\right)^{-1} \left(\mbA+\mbC\right): \nabla\mbg\right]\\
    &= \epsilon\mathbb{E}_{\mbtheta\sim\rho} \left[\left(\mbI + \nabla((\mbA(\mbtheta)-\mbC(\mbtheta))\mbf(\mbtheta))\right)^{-1} \left(\mbA(\mbtheta)+\mbC(\mbtheta)\right): \langle\nabla_1 k(\mbtheta,\cdot), \mbg(\cdot)\rangle_{\mathcal{H}^D}\right]\\
    &= \epsilon\bigg\langle\mathbb{E}_{\mbtheta\sim\rho} \left[\left(\mbI + \nabla((\mbA(\mbtheta)-\mbC(\mbtheta))\mbf(\mbtheta))\right)^{-1} \left(\mbA(\mbtheta)+\mbC(\mbtheta)\right)\nabla_1 k(\mbtheta,\cdot), \mbg(\cdot)\right]\bigg\rangle_{\mathcal{H}^D},\\
    \Delta_4 &= \epsilon\mathbb{E}_{\mbtheta\sim\rho} \left[\underbrace{\left(\mbI + \nabla((\mbA(\mbtheta)-\mbC(\mbtheta))\mbf(\mbtheta))\right)^{-1}}_{\mbQ(\mbtheta)} : \mbM\right]\\
    &= \epsilon\mathbb{E}_{\mbtheta\sim\rho}\sum_{i,j=1}^D \mbQ_{ij}\mbM_{ij} = \epsilon \big\langle\mathbb{E}_{\mbtheta\sim\rho}\sum_{i,j=1}^D\mbQ_{ij}\mbh^{(ij)}k(\mbtheta,\cdot), \mbg(\cdot)\big\rangle_{\mathcal{H}^D}
\end{align}
It is straightforward that when $\mbf = \mathbf{0}$,
\begin{align}
    \Delta_4 &= \epsilon\mathbb{E}_{\mbtheta\sim\rho} \big\langle\sum_{i=1}^D \mbh^{(ii)}k(\mbtheta,\cdot), \mbg\big\rangle_{\mathcal{H}^D}, \\
    \sum_{i=1}^D \mbh_\ell^{(ii)} &= \sum_{i=1}^D \frac{\partial (\mbA-\mbC)_{i\ell}}{\partial\mbtheta_i}\\
    &= \sum_{i=1}^D \frac{\partial (\mbA+\mbC)_{\ell i}}{\partial\mbtheta_i} = \nabla \cdot(\mbA+\mbC).
\end{align}
Using the definition of $\Delta_i, i\in[4]$, we can derive the GSVGD particle update as $\mbf = \mathbf{0}$, confirming the functional derivative coinciding with the GSVGD particle update.
\subsection{Projection onto RKHS}
To prove $\mbv_\mathcal{H}^{\mbA,\mbC}$ is the projection of $\mbv^{\mbA,\mbC}$ onto $\mathcal{H}^D$, we start from the inner product on $\mathcal{L}_\rho^2$, such that $\forall \mbv\in\mathcal{H}^D$
\begin{align}
    \langle\mbv^{\mbA,\mbC}(\cdot\vert\rho), \mbv\rangle_{\mathcal{L}_{\rho}^2} &= \mathbb{E}_\rho \left[(\mbA+\mbC)\nabla\log \pi/\rho \cdot \mbv\right]\\
    &= \mathbb{E}_\rho\left[(\mbA+\mbC)\nabla\log \pi \cdot \mbv\right] - \mathbb{E}_\rho\left[(\mbA+\mbC)\nabla\log \rho \cdot \mbv\right]\\
    &= \mathbb{E}_\rho\left[(\mbA+\mbC)\nabla\log \pi \cdot \mbv\right] - \underbrace{\int\left[(\mbA+\mbC)\nabla\rho \cdot \mbv\right]\text{d}\mbtheta}_{\text{weak derivative of measures}}\\
    &= \mathbb{E}_\rho\left[(\mbA+\mbC)\nabla\log \pi \cdot \mbv\right] + \mathbb{E}_\rho \left[\sum_{i, j} \frac{\partial((\mbA_{ij}+\mbC_{ij})v_i)}{\partial \mbtheta_j}\right]\\
    &= \mathbb{E}_\rho\left[\left\{(\mbA+\mbC)\nabla\log \pi + \nabla\cdot(\mbA+\mbC)\right\} \cdot \mbv\right] + \mathbb{E}_\rho \left[\sum_{i, j} \frac{(\mbA_{ij}+\mbC_{ij})\partial v_i}{\partial \mbtheta_j}\right]\\
    &= \big\langle\mathbb{E}_{\mbtheta\sim\rho}\left\{(\mbA(\mbtheta)+\mbC(\mbtheta))\nabla\log \pi(\mbtheta) + \nabla\cdot(\mbA(\mbtheta)+\mbC(\mbtheta))\right\} k(\mbtheta,\cdot), \mbv(\cdot)\big\rangle_{\mathcal{H}^D}\\
    &+ \mathbb{E}_\rho \left[\sum_{i, j} \frac{(\mbA_{ij}+\mbC_{ij})\partial v_i}{\partial \mbtheta_j}\right],
\end{align}
\begin{align}
    \mathbb{E}_\rho \left[\sum_{i, j} \frac{(\mbA_{ij}+\mbC_{ij})\partial v_i}{\partial \mbtheta_j}\right] &= \mathbb{E}_{\mbtheta\sim\rho} \left[\sum_{i, j} (\mbA_{ij}(\mbtheta)+\mbC_{ij}(\mbtheta))\frac{\partial v_i(\mbtheta)} {\partial\mbtheta_j}\right]\\
    &= \mathbb{E}_{\mbtheta\sim\rho} \left[\sum_{i, j} (\mbA_{ij}(\mbtheta)+\mbC_{ij}(\mbtheta))\big\langle\frac{\partial}{\partial\mbtheta_j} k(\mbtheta, \cdot), \mbv_i(\cdot)\big\rangle_{\mathcal{H}}\right]\\
    &= \bigg\langle\mathbb{E}_{\mbtheta\sim\rho}  \left[(\mbA+\mbC)\nabla k(\mbtheta, \cdot)\right], \mbv(\cdot)\bigg\rangle_{\mathcal{H}^D}.
\end{align}
Therefore, confirming that $\forall \mbv\in\mathcal{H}^D, \langle \mbv^{\mbA,\mbC}, \mbv\rangle_{\mathcal{L}_\rho^2} = \langle \mbv_{\mathcal{H}}^{\mbA,\mbC}, \mbv\rangle_{\mathcal{H}^D}$.
\subsection[a]{Interpreting GSVGD as MCMC dynamics of $\pi^{\otimes N}$}
Additionally, we can view GSVGD with with constant $\mbC$ matrices and $N$ particles as the mean-field limit of a MCMC dynamics inferring the product target measure $\underbrace{\pi\times\hdots\times\pi}_{N} = \pi^{\otimes N}$, a rather trivial extension of the discussion in \citep{gallego2018stochastic}. The $\mbA(\mbtheta^{\otimes N}), \mbC(\mbtheta^{\otimes N}) \in \mathbb{R}^{ND\times ND}$ is defined as 
\begin{align}
    \tilde{\mbA}_{i, j} = \frac{1}{N} k(\mbtheta_i, \mbtheta_j) \frac{\mbA(\mbtheta_i) + \mbA(\mbtheta_j)}{2},\\
    \tilde{\mbC}_{i, j} = \frac{1}{N} k(\mbtheta_i, \mbtheta_j) \left[\mbC + \frac{\mbA(\mbtheta_j) - \mbA(\mbtheta_i)}{2}\right].
\end{align}
We can verify that the MCMC dynamics
\begin{align}
    \dot{\mbtheta}_t^{\otimes N} &= \frac{1}{\pi(\mbtheta^{\otimes N})}\nabla\cdot\left(\pi(\mbtheta_t^{\otimes N})\left(\tilde{\mbA}(\mbtheta_t^{\otimes N})+\tilde{\mbC}(\mbtheta_t^{\otimes N})\right)\right) + \sqrt{2\tilde{\mbA}(\mbtheta_t^{\otimes N})}\text{d}\mbW_{ND},
\end{align}
takes the invariant measure $\pi^{\otimes N}$. And that the drift coefficient corresponds to the GSVGD particle update. Furthermore, this framework accepts non-constant $\mbC$ matrices when $\tilde{\mbA}_{i,j} = \frac{1}{N} k(\mbtheta_i, \mbtheta_j) \left[\frac{\mbA(\mbtheta_i) + \mbA(\mbtheta_j)}{2} + \frac{\mbC(\mbtheta_j) - \mbC(\mbtheta_i)}{2}\right]$ remains positive semidefinite. \par
It is worth noting that as $N\rightarrow \infty$, the drift coefficient goes to $\mathbf{0}$, making GSVGD the mean-field limit of such dynamics.
\subsection{Stochastic particle optimization sampling (SPOS) as MCMC dynamics}
Viewing GSVGD as the mean-field limit of MCMC dynamics yields additional insights. For example, \citet{zhang2020stochastic} propose stochastic particle optimization sampling in the form of 
\begin{align}
    \dot{\mbtheta}_t^{\otimes N} &= \frac{1}{\pi(\mbtheta_t^\otimes N)}\nabla\cdot\left(\pi(\mbtheta_t^{\otimes N})\left(\overline{\mbK}\otimes \mbI + \sigma^2 \mbI\right)\right) + \sqrt{2\sigma^2\mbI} \mbW_{ND},
\end{align}
where $\overline{\mbK}_{ij} = k(\mbtheta_i, \mbtheta_j)$. 
Formally, SPOS combines the SVGD particle update with a step of (multi-chained) Langevin diffusion. The SPOS particle update does not conform to the standard formulation of MCMC dynamics, yielding a biased sampling algorithm. However, such bias can be fixed by changing the diffusion coefficient into $\sqrt{2\left(\overline{\mbK}\otimes \mbI+\sigma^2\mbI\right)}$. Translating into discretized dynamics, SPOS generates samples from the target measure when the injected noise is correlated across particles. However, the correlated injected noise has variance approaching zero as $N\rightarrow\infty$. 
\subsection{Recovering SVGD with GFSF}
While SVGD takes the form of gradient flow on $(\mathcal{P}(\Omega), W_{\mathcal{H}})$, we can connect SVGD with other form of smoothing discussed in \citet{liu2019understanding}, noted as gradient flow with smoothed test functions (GFSF). GSVGD taking $\mbA = \overline{\mbK}\otimes \mbI, \mbC = \mathbf{0}$ gives 
\begin{align}
    \mbv^{\mbA,\mbC} &= \overline{\mbK}\otimes \mbI \nabla\log\pi^{\otimes N}/\rho^{\otimes N} = \overline{\mbK}\otimes \mbI \nabla\log\pi^{\otimes N} - \overline{\mbK}\otimes \mbI \nabla\log\rho^{\otimes N}.
\end{align}
Using the GFSF estimation of $\nabla\log\rho$ \footnote{formally, GFSF is equivalent of applying the Stein gradient estimator \citep{li2017gradient} without regularization.}, the term $\overline{\mbK}\otimes \mbI \nabla\log\rho^{\otimes N}$ yields $\nabla\cdot \overline{\mbK}$, recovering the SVGD particle update. Viewing PARVI in the product space yields additional insights of connecting different smoothing methods. Analogously, we can use the generalized Stein's identity \citep{gorham2019measuring} to arrive at a GFSF smoothing of GSVGD. 
\section{Additional discussion}
\subsection[b]{Formulating Riemannian Langevin diffusion as gradient flow on $(\mathcal{P}(\Omega), W_{2, \mbA})$}
With $\mbC=\mathbf{0}$ and positive definite $\mbA$, we can generalize the 2-Wasserstein metric in the Benamou-Brenier form \citep{benamou2000a}
\begin{align}
    W_{2, \mbA}^2(\rho_0, \rho_1) = \inf_{\phi,\rho_t} &\left\{\int_0^1 \int \langle\nabla\phi_t, \mbA\nabla\phi_t\rangle\dx{t}: \dot{\rho}_t + \nabla \cdot (\rho_t \mbA\nabla\phi_t) = 0\right\}.
\end{align}
The Onsager operator of $W_{2, \mbA}$ takes the form $G(\rho)^{-1}: \phi \mapsto -\nabla\cdot\left(\rho \mbA \nabla\phi\right)$. MCMC dynamics such as Riemannian Langevin diffusion \citep{girolami2011riemann} can be interpreted as a gradient flow of $\KL{\rho}{\pi}$ on $W_{2, \mbA}$: $\dot{\rho}_t=\nabla\cdot\left(\rho_t\mbA\nabla\frac{\delta\KL{\rho_t}{\pi}}{\delta\rho_t}\right) = -G(\rho_t)^{-1}\frac{\delta\KL{\rho_t}{\pi}}{\delta\rho_t}$, circumventing the necessity of defining a projection $\mathfrak{p}_\rho$ onto the tangent space of $(\mathcal{P}(\Omega), W_2)$. 
\subsection{How to accelerate PARVI for underdamped Langevin diffusion?}
\citet{ma2019is} argue that the underdamped Langevin diffusion with $\mbA = \left(\begin{matrix}\mathbf{0}&\mathbf{0}\\ \mathbf{0}& A\mbI\end{matrix}\right), \mbC = \left(\begin{matrix}\mathbf{0} & -\mbI\\ \mbI & \mathbf{0}\end{matrix}\right)$ is an analog of Nesterov's acceleration of the overdamped Langevin diffusion $\mbA = \mbI, \mbC = \mathbf{0}$, and such analog still stands for their PARVI variants. Similar to results presented in MCMC research \citep{mou2021high}, we can construct PARVI with third-order Langevin diffusion $\mbA = \left(\begin{matrix}\mathbf{0}&\mathbf{0} &\mathbf{0}\\ \mathbf{0}&\mathbf{0} &\mathbf{0}\\ \mathbf{0}&\mathbf{0}& A\mbI\end{matrix}\right)$, $\mbC =\left(\begin{matrix}\mathbf{0} & -\mbI & \mathbf{0}\\ \mbI & \mathbf{0} & -\gamma\mbI\\ \mathbf{0} & \gamma\mbI & \mathbf{0}\end{matrix}\right)$, equivalent to applying a higher-order momentum method in gradient descent. 
\subsection{Momentum resampling}
One unexplored aspect of particle variational inference with momentum variable is the possibility of turning PARVI into a proper sampling algorithm, one feat unattainable by de-randomization of LD, as deterministic optimization of $N$ particle can only produce $N$ samples. With the introduction of momentum variables, it is possible to periodically resample the momentum variable to obtain more samples from the target distribution -- in practice, it involves combining the deterministic particle updates with a jump process that routinely samples from the marginal distribution of momentum.\par
Aside from possibly obtaining more samples, resampling during the optimization can also speed up convergence to the target distribution. As we know the marginal distribution with respect to $\mbr$, resampling of momentum variables reduces the KL-divergence between $\rho_t$ and $\pi$, as $\KL{\rho_t(\theta)\pi(\mbr)}{\pi(\mbtheta)\pi(\mbr)} \leq \KL{\rho_t(\theta,\mbr)}{\pi(\mbtheta)\pi(\mbr)}$. It remains a theoretical and empirical open question whether resampling momentum can speed up convergence.
\section{Experiment details}
\subsection{Toy experiments}
The 2-dimensional likelihood of the toy experiments used in this paper takes the form of $\pi \propto \frac{1}{3}\sum_{i=1}^3 \exp\left(-\frac{x^4}{10} + \frac{(z_iy-x^2)^2}{2}\right)$, $z_i = \{-2, 0, 2\}$, and the original particle locations are initialized as $\left(\begin{matrix} x\\ y\end{matrix}\right)\sim \mathcal{N}(0, 0.01\mbI)$. We can use the energy function $U(\mbtheta) = \log\left(\frac{1}{3}\sum_{i=1}^3 \exp\left(-\frac{x^4}{10} + \frac{(z_iy-x^2)^2}{2}\right)\right)$ to construct Riemannian samplers in the experimental setting consistent with \citet{ma2015a}, where Fisher information metric matrix is defined as $\mbG^{-1}(\mbtheta)=D\sqrt{|U(\mbtheta)+C|}, U=1.5, C=0.5$. In Riemannian \gls{SVGD} \citep{liu2017riemannian}, we follow the practice of Riemannian LD \citep{girolami2011riemann} and parametrize $\mbA(\mbtheta) = \mbG^{-1}(\mbtheta), \mbC(\mbtheta)=\mathbf{0}$; In Riemannian SGHMC, we follow \citep{ma2015a} parametrize $\mbA(\mbtheta) = \left(\begin{matrix}\mathbf{0}&\mathbf{0}\\ \mathbf{0}&\mbG^{-1}(\mbtheta)\end{matrix}\right), \mbC(\mbtheta) = \left(\begin{matrix}\mathbf{0} & -\mbG^{-1/2}\\ \mbG^{-1/2} & \mathbf{0}(\mbtheta)\end{matrix}\right)$.
\subsection{Bayesian neural network experiments}
In Bayesian neural network for regression, we use a standard structure of $1$ hidden layers with width $50$, along with a conjugate prior on the precision parameter of its weight priors. Specifically, we have
\begin{align}
    y &\sim \mathcal{N}\left(\mbW_2^{\top}\text{ReLU}(\mbW_1^\top\mbx + \mbb_1) + \mbb_2, \gamma^{-1}\right),\\
    \mbW_1, \mbb_1, \mbw_2, \mbb_2 &\sim \mathcal{N}(\mathbf{0}, \lambda^{-1}\mbI), \mbW_1 \in \mathbb{R}^{D\times 50}, \mbb_1\in\mathbb{R}^{50},\\
    \gamma, \lambda &\sim \text{Gamma}(1, 0.1).
\end{align}
The weights $\mbW_i$ are initialized with glorot normal distribution and $\mbb_i$ are intialized with zero. \par
The hyperparameters for the methods applied in the paper are selected by cross-validation in the following fashion: the learning rate $\eta$ is selected in $\eta\in\{10^{-8}, 10^{-7}, 10^{-6}, 10^{-5}, 10^{-4}, 10^{-3}, 10^{-2}\}$; for methods involving additional momentum variables $\mbr$, we adopt the momentum interpretation of the underdamped Langevin dynamics \citep{chen2014stochastic} and select momentum term $\alpha \in \{0.01, 0.1, 0.5\}$ \footnote{It is notable that the step size $\epsilon$ in the discretization of dynamics does not directly correspond to the ``effective learning rate'' in SGHMC-type samplers. }; for thermostat-type samplers, we additionally tune the precision parameter of the temperature variable, with $\mu\in\{0.1, 1.0, 10.0\}$. For methods involving kernel parameters, we parameterize a squared exponential kernel with the median method \citep{liu2016stein}. We take symmetric splitting for methods involving momentum variables. \par
For the 6 datasets from UCI repository, we take a $90\% / 10\%$ training/test partition of the data; for the 6 medium-sized datasets (except for ``year'' and ``protein''), we take 20 different training / test splits, 6 splits for ``protein'', and 6 different initializations with the ``year'' dataset, as the split is fixed. We implemented the models using JAX \citep{jax2018github}, and ran the experiments on Nvidia Volta V100 GPU nodes. We present the technical formulation of the methods and its running time in Table \ref{tab:config} and Table \ref{tab:run_time}, respectively.
\begin{table}[ht!]
    \centering
    \resizebox{\textwidth}{!}{%
    \begin{tabular}{l|cccc|r}
    \toprule
    Method & $\mbA$ & $\mbC$ & $\pi$ & hyperparameters & Reference\\
    \midrule
        LD & $\mbI$ & $\mathbf{0}$ & $\pi(\mbtheta)$ & $\eta=\epsilon$ & \citet{welling2011bayesian}\\
        SVGD & $\mbI$ & $\mathbf{0}$ & $\pi(\mbtheta)$ &$\eta=\epsilon$ & \citet{liu2016stein}\\
        Blob & $\mbI$ & $\mathbf{0}$ & $\pi(\mbtheta)$ & $\eta=\epsilon$ & \citet{chen2018a}\\
        DE & - & - & $\pi(\mbtheta)$ & $\eta=\epsilon, \alpha$ & \citet{lakshminarayanan2017simple}\\
        SGHMC-Blob & $\left(\begin{matrix}\mathbf{0} & \mathbf{0} \\ \mathbf{0} & A\mbI\end{matrix}\right)$ & $\left(\begin{matrix}\mathbf{0} & -\mbI \\ \mbI & \mathbf{0}\end{matrix}\right)$ & $\pi(\mbtheta)\mathcal{N}(\mbr\vert\mathbf{0}, \sigma^2\mbI)$ & $\eta = \epsilon^2 \sigma^{-2}, \alpha = \epsilon\sigma^{-2} A$ & \citet{liu2019mcmc}\\
        SGHMC-Stein & $\left(\begin{matrix}\mathbf{0} & \mathbf{0} \\ \mathbf{0} & A\mbI\end{matrix}\right)$ & $\left(\begin{matrix}\mathbf{0} & -\mbI \\ \mbI & \mathbf{0}\end{matrix}\right)$ & $\pi(\mbtheta)\mathcal{N}(\mbr\vert\mathbf{0}, \sigma^2\mbI)$ & $\eta = \epsilon^2 \sigma^{-2}, \alpha = \epsilon\sigma^{-2} A$ & this work \\
        SGNHT & $\left(\begin{matrix}\mathbf{0} & \mathbf{0} & \mathbf{0}\\ \mathbf{0} & A\mbI & \mathbf{0}\\ \mathbf{0} & \mathbf{0} & \mathbf{0}\end{matrix}\right)$ & $\left(\begin{matrix}\mathbf{0} & -\mbI & \mathbf{0}\\ \mbI & \mathbf{0} & (\mu\sigma^2)^{-1}\text{diag}(\mbr)\\ \mathbf{0} & -(\mu\sigma^2)^{-1}\text{diag}(\mbr) & \mathbf{0}\end{matrix}\right)$ & $\pi(\mbtheta)\mathcal{N}(\mbr\vert\mathbf{0}, \sigma^2\mbI)\mathcal{N}(\mbxi\vert A\mathbf{1}, \mu^{-1}\mbI)$ & $\eta = \epsilon^2 \sigma^{-2}, \alpha = \epsilon\sigma^{-2} A, \mu$ & \citet{ding2014bayesian}\\
        SGNHT-Stein & $\left(\begin{matrix}\mathbf{0} & \mathbf{0} & \mathbf{0}\\ \mathbf{0} & A\mbI & \mathbf{0}\\ \mathbf{0} & \mathbf{0} & \mathbf{0}\end{matrix}\right)$ & $\left(\begin{matrix}\mathbf{0} & -\mbI & \mathbf{0}\\ \mbI & \mathbf{0} & (\mu\sigma^2)^{-1}\text{diag}(\mbr)\\ \mathbf{0} & -(\mu\sigma^2)^{-1}\text{diag}(\mbr) & \mathbf{0}\end{matrix}\right)$ & $\pi(\mbtheta)\mathcal{N}(\mbr\vert\mathbf{0}, \sigma^2\mbI)\mathcal{N}(\mbxi\vert A\mathbf{1}, \mu^{-1}\mbI)$ & $\eta = \epsilon^2 \sigma^{-2}, \alpha = \epsilon\sigma^{-2} A, \mu$ & this work\\
        \bottomrule
    \end{tabular}
    }
    \caption{An overview of the parameters in the methods used in the BNN experiments paper, including the (possibly augmented) target distribution, the parameterizations of MCMC dynamics, and the tunable hyperparameters (step size $\eta$, momentum term $\alpha$ and precision term for the temperature variable $\mu$).}
    \label{tab:config}
\end{table}

\begin{table}[ht!]
    \centering
    \resizebox{\textwidth}{!}{%
    \begin{tabular}{l|cccccccc}
\toprule
{} &         boston &       concrete &        energy &         kin8nm &          power &          yacht &             year &         protein \\
\midrule
LD          &  124.88(22.68) &   91.40(45.34) &  98.17(41.45) &  131.89(21.67) &  131.01(14.20) &   86.91(42.83) &   811.08(140.66) &    559.76(9.84) \\
SVGD        &   75.76(10.04) &   75.31(13.82) &  67.70(19.69) &    87.03(3.68) &   81.98(10.59) &   71.11(10.14) &    704.39(30.75) &   348.16(16.67) \\
Blob        &    79.69(2.25) &   75.39(14.36) &  66.79(19.86) &   83.55(10.36) &   75.35(18.10) &   66.94(18.52) &    708.06(16.48) &   352.89(11.15) \\
HMC-Blob    &   76.76(20.62) &   80.95(19.51) &  83.31(16.53) &   93.61(13.63) &   87.99(16.13) &   73.33(22.05) &   874.68(165.39) &   409.75(13.78) \\
SGNHT       &  231.86(55.61) &  245.63(25.54) &  252.53(3.79) &   256.84(2.74) &  232.18(60.56) &  222.90(59.43) &  1326.01(197.56) &  1016.57(79.80) \\
DE          &   76.76(10.52) &   75.53(14.57) &   75.30(7.04) &   80.76(13.80) &   79.26(15.56) &   70.80(13.89) &    709.06(23.05) &   351.96(13.21) \\
SGHMC-Stein &   89.03(10.60) &    91.82(8.58) &  85.50(13.45) &    95.02(9.94) &    90.87(9.73) &   81.02(16.11) &   761.57(119.60) &   395.49(14.61) \\
SGNHT-Stein &   101.22(3.27) &   94.33(13.16) &  89.52(18.69) &   86.18(19.88) &   96.86(14.21) &   89.98(13.62) &    879.78(14.91) &   429.61(13.22) \\
\bottomrule
\end{tabular}
}
    \caption{Mean (standard deviation) of running times for BNN experiments measured in seconds. All methods are run for 5,000 iterations (first 6 columns, averaged over 20 runs) and 20,000 iterations (last 2 columns, averaged over 6 runs), respectively.}
    \label{tab:run_time}
\end{table}
\subsection{Additional experiments}
Apart from the standard from of GSVGD and Blob methods in the paper, we experimented Bayesian neural network with \gls{PARVI} consistent with the pSGHMC-det formula in \citet{liu2019mcmc}, and the corresponding Stein version with \gls{RKHS} projection. The experiment results do not show clear difference from the standard form. From the perspective of Hamiltonian dynamics, we can view underdamped \gls{LD} and this variant of \gls{PARVI} both as Hamiltonian Monte Carlo with a ``continuous resampling'' of the momentum variable: underdamped \gls{LD} resamples momentum by running overdamped \gls{LD} on the momentum variable; its \gls{PARVI} variant runs \gls{SVGD} (Stein) or Blob variant as continuous resampling. While the particles do not converge to an equilibrium, the marginal distribution $\rho_t$ remains unchanged. 
\begin{table}[ht!]
    \centering
    \resizebox{\textwidth}{!}{%
    \begin{tabular}{l|cccccccc}
\toprule
{} &         boston &       concrete &        energy &         kin8nm &          power &          yacht &             year &         protein \\
\midrule
SGHMC-Blob*	& 11.68(39.93)	& 0.21(0.45)	& 0.00(0.00)	& 0.07(0.00)	& 0.05(0.00) & 	0.00(0.00)&	0.64(0.00)&	0.49(0.01)\\
SGHMC-Stein* & 	0.12(0.07)& 	0.09(0.02)& 	0.00(0.00)&	0.07(0.00)&	0.05(0.00)&	0.00(0.00)&	0.65(0.00)&	0.48(0.01)\\
\bottomrule
    \end{tabular}
    }
    \caption{Mean (standard deviation) of mean squared error with \gls{PARVI} experiment: the experimental setting is the same as the standard BNN experiment. }
\end{table}
\begin{table}[ht!]
    \centering
    \resizebox{\textwidth}{!}{%
    \begin{tabular}{l|cccccccc}
\toprule
{} &         boston &       concrete &        energy &         kin8nm &          power &          yacht &             year &         protein \\
\midrule
SGHMC-Blob*	& -2.58(0.12)&	-2.93(0.08)&	-0.49(0.10)&	1.23(0.02)&	-2.77(0.04)&	-0.70(0.36)&	-3.58(0.00)&	-2.87(0.01)\\
SGHMC-Stein* & 	-2.54(0.28)	& -2.98(0.09)&	-0.32(0.24)&	1.25(0.02)&	-2.78(0.03)&	-0.76(0.53)&	-3.59(0.00)&	-2.87(0.01)\\
\bottomrule
    \end{tabular}
    }
    \caption{Mean (standard deviation) of text log-likelihood with \gls{PARVI} experiment: the experimental setting is the same as the standard BNN experiment. }
\end{table}
\newpage
\bibliography{refs}
\bibliographystyle{unsrtnat}